\documentclass[letterpaper]{article} 
\usepackage{aaai24}  
\usepackage{times}  
\usepackage{helvet}  
\usepackage{courier}  
\usepackage[hyphens]{url}  
\usepackage{graphicx} 
\usepackage{natbib}  
\usepackage{caption} 
\usepackage{algorithm}
\usepackage{algorithmic}

\usepackage{newfloat}
\usepackage{listings}
\DeclareCaptionStyle{ruled}{labelfont=normalfont,labelsep=colon,strut=off} 
\floatstyle{ruled}
\newfloat{listing}{tb}{lst}{}
\floatname{listing}{Listing}
\usepackage{amsmath}
\usepackage{booktabs}    
\usepackage{bm}
\usepackage{graphicx}
\usepackage{subfigure}
\usepackage{graphicx}

\title{UV-SAM: Adapting Segment Anything Model for Urban Village Identification}

\author{
    Xin Zhang\textsuperscript{\rm 1},
    Yu Liu\textsuperscript{\rm 2}\thanks{Corresponding author},
    Yuming Lin\textsuperscript{\rm 2},
    Qingmin Liao\textsuperscript{\rm 1},
    Yong Li\textsuperscript{\rm 2} 
}
\affiliations{
    \textsuperscript{\rm 1}Shenzhen International Graduate School, Tsinghua University, Shenzhen, China\\
    \textsuperscript{\rm 2}Department of Electronic Engineering, Tsinghua University, Beijing, China\\
    zhangxin4087@163.com, liuyu2419@126.com
}

\usepackage{bibentry}

\begin{document}

\maketitle

\begin{abstract}
Urban villages, defined as informal residential areas in or around urban centers, are characterized by inadequate infrastructures and poor living conditions, closely related to the Sustainable Development Goals (SDGs) on poverty, adequate housing, and sustainable cities. Traditionally, governments heavily depend on field survey methods to monitor the urban villages, which however are time-consuming, labor-intensive, and possibly delayed. Thanks to widely available and timely updated satellite images, recent studies develop computer vision techniques to detect urban villages efficiently. However, existing studies either focus on simple urban village image classification or fail to provide accurate boundary information. To accurately identify urban village boundaries from satellite images, we harness the power of the vision foundation model and adapt the Segment Anything Model (SAM) to urban village segmentation, named UV-SAM. Specifically, UV-SAM first leverages a small-sized semantic segmentation model to produce mixed prompts for urban villages, including mask, bounding box, and image representations, which are then fed into SAM for fine-grained boundary identification. Extensive experimental results on two datasets in China demonstrate that UV-SAM outperforms existing baselines, and identification results over multiple years show that both the number and area of urban villages are decreasing over time, providing deeper insights into the development trends of urban villages and sheds light on the vision foundation models for sustainable cities. The dataset and codes of this study are available at \url{https://github.com/tsinghua-fib-lab/UV-SAM}.
\end{abstract}

\section{Introduction}

\begin{figure}[tbp]
    \centering
    \subfigure[Narrow alleyways]{
        \includegraphics[width=0.226\textwidth]{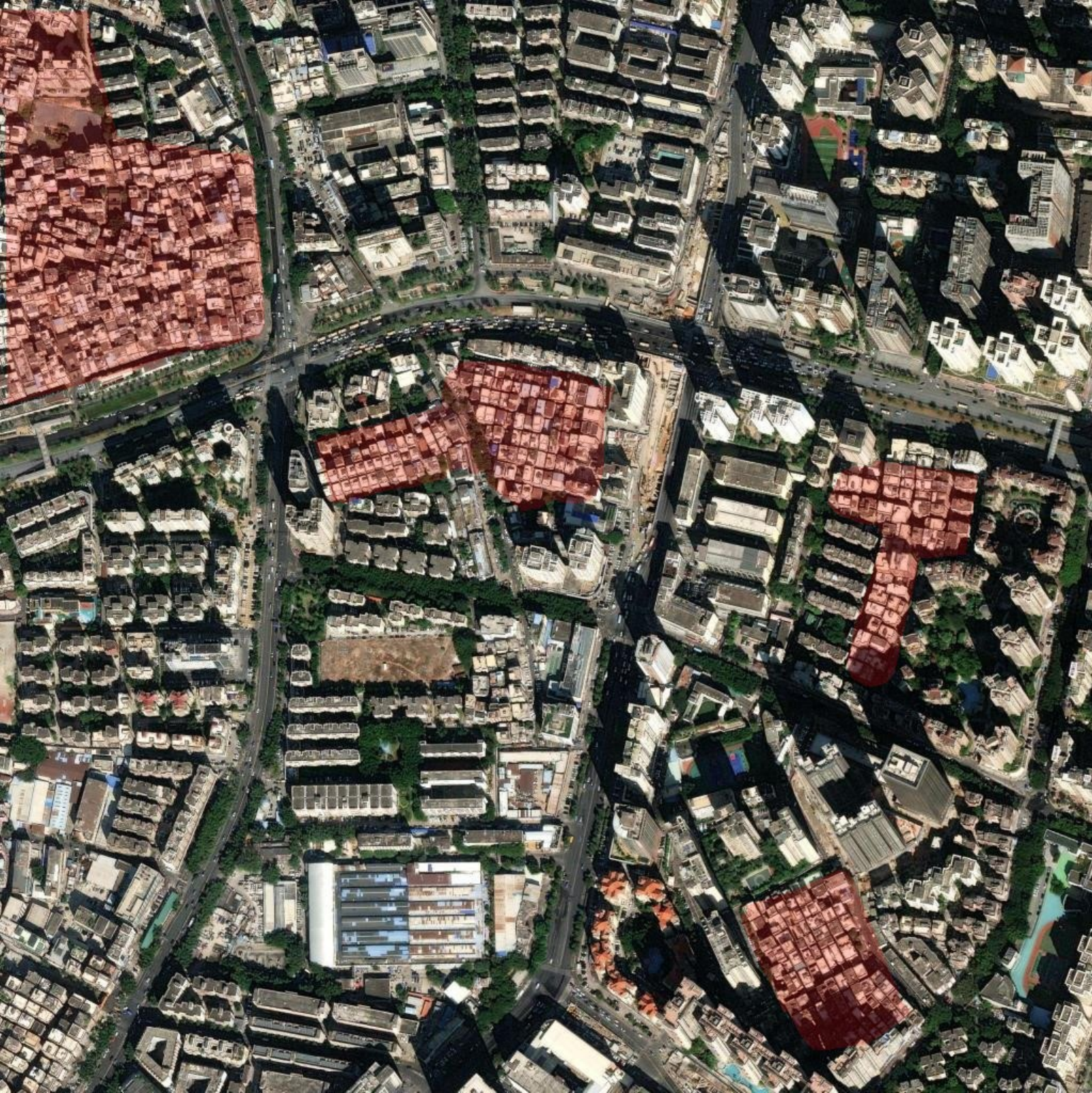}
    }
    \subfigure[Older low-rise buildings]{
    \includegraphics[width=0.226\textwidth]{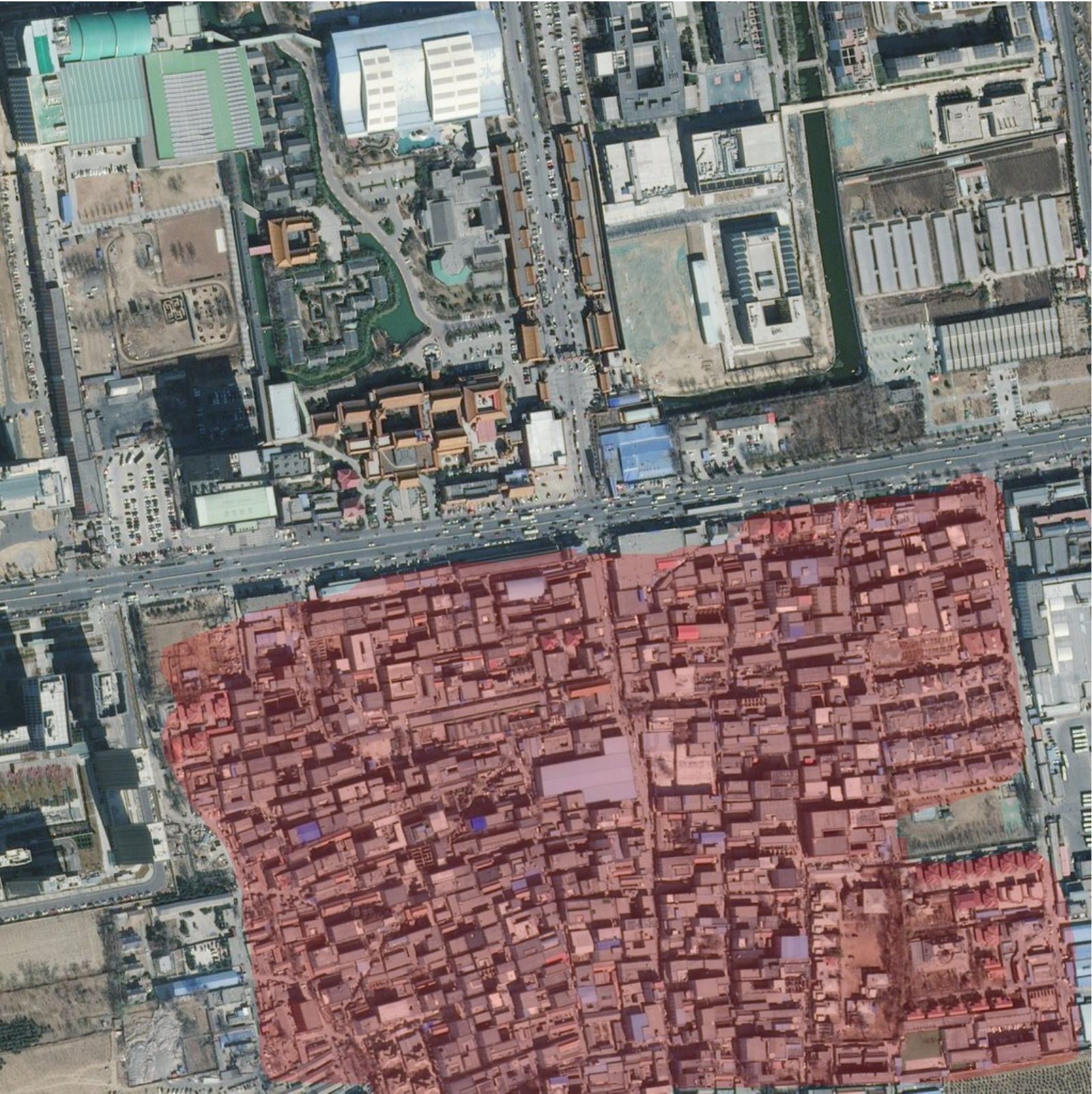}}
    \caption{Examples of urban villages identified from satellite images, with appearance characteristics provided. The red part represents the urban village areas.}
    \label{fig:uv_eg}
\end{figure}

As a representative type of informal settlement, urban villages are densely populated neighborhoods in both the outskirts and the downtown segments of major Chinese cities, typically consisting of older low-rise buildings and narrow alleyways \cite{enwiki:1188353043}, as shown in Figure~\ref{fig:uv_eg}. On the one hand, urban villages provide affordable housing options for migrant workers and low-income citizens, contributing to the socioeconomic fabric of cities. On the other hand, urban villages often face challenges related to inadequate infrastructure, limited access to public services, and poor living conditions \cite{chen2021uvlens}. Hence, aligning well with United Nations' 11th Sustainable Development Goal (SDG 11)``Making cities and human settlements inclusive, safe, resilient and sustainable" \cite{nath2016making}, accurately identifying urban villages is essential for both urban planning and governance in future sustainable cities.

Traditionally, urban village identification heavily depends on field surveys and manual mapping \cite{zheng2009urban}, where urban planners would visit different areas, collect socioeconomic data, and visually identify urban village boundaries. While such methods provide valuable insights, they are time-consuming, labor-intensive, and limited in spatiotemporal coverage. In recent years, exploring computer vision techniques with satellite images for urban villages has gained significant attention. Most studies build image classification models to classify whether a given satellite image contains an urban village \cite{chen2022multi,fan2022multilevel,fan2022urban,xiao2023contextual} without boundaries identified, while others explore semantic segmentation models to identify urban village boundaries in satellite images \cite{mast2020mapping,pan2020deep,chen2019identifying}. However, due to the complex background interference in satellite images and the lack of well-defined boundaries between urban villages and surrounding neighborhoods, existing studies perform poorly in providing accurate urban village boundaries, which further hinders the estimation of the areas and expansions of urban villages \cite{kirillov2023segment}. Moreover, limited labeled data in urban villages also make segmentation models prone to overfit and fail to generalize to noisy satellite images, e.g., occlusion and seasons. 

Meanwhile, owing to training on over one billion images, the recent vision foundation model of the Segment Anything Model (SAM) exhibits remarkable generalization capabilities as well as high mask quality for category-agnostic segmentation, which is quite sensible to segment boundaries \cite{kirillov2023segment} and has been investigated into various domains. Specifically, SAM operates in a manner that requires a preexisting prompt such as a reference point, bounding box, or mask, to accompany the input image. Obviously, category-agnostic segmentation provided by SAM cannot be directly applied to semantic segmentation. Hence, several studies explore refined manual prompts for category-specific segmentation in domain-specific applications, such as manually-labeled bounding boxes for medical image segmentation \cite{wu2023medical}, showcasing promising results. Therefore, considering the limitation of blurry boundary recognition in existing urban village identification studies as well as the strength of generalization and boundary sensitivity in SAM, an interesting research question is whether SAM can help urban village identification from satellite images. 

Regarding the research question above, in this paper, we propose a generalist-specialist-like framework called UV-SAM, to adapt SAM for urban village identification. Specifically, the critical point of adaption lies in generating category-specific prompts that can encourage SAM to focus on urban villages in satellite images. Therefore, we regard SAM with large frozen parameters as a generalist in category-agnostic segmentation and develop a semantic segmentation model with limited learnable parameters as a specialist for urban village identification, where the specialist automatically generates prompts for the generalist while the outputs by the generalist in turn update model parameters of the specialist. Following the proposed framework, UV-SAM employs four distinct categories of prompts specifically for urban villages in satellite images. Firstly, UV-SAM develops a small-sized semantic segmentation model like SegFormer \cite{xie2021segformer} to produce coarse segmentation masks for urban villages, based on which mask prompts and box prompts of urban villages are generated. Secondly, the feature maps from image encoders in both SAM and SegFormer are extracted as semantic prompts. Furthermore, a prompt mixer module is designed to fuse such four types of prompts together, and the resulting urban village prompt vector is fed into SAM for urban village specific segmentation. In summary, our contributions lie in three aspects:

\begin{itemize}
    \item We are the first to introduce the vision foundation model SAM for urban village identification, which enlightens the application of foundation models in artificial intelligence for sustainable cities and SDG.

    \item We establish a novel generalist-specialist framework, UV-SAM, which automatically generates four distinct types of prompts, and seamlessly integrates SAM into urban village identification applications.
    
    \item We conduct extensive experiments on two cities Beijing and Xi'an in China, and the results demonstrate that our proposed framework achieves significant performance improvement compared with state-of-the-art models. Further case studies reveal the evolving trends of urban villages in both amount and area, as well as their spatial distribution, which provides valuable insights for urban planning and governance.
\end{itemize}

\section{Related Works}

\subsubsection{Satellite Image-based Urban Village Identification.}

Urban village identification refers to the process of identifying areas or regions within a city that exhibit characteristics of urban villages, which are crucial for understanding the spatial distribution and evolution of urban villages.

Several studies investigate the satellite image classification problem to identify whether urban villages exist in corresponding images. Earlier studies \cite{huang2015spatiotemporal,liu2017use} apply traditional machine learning algorithms, such as support vector machines, to classify urban and non-urban areas based on handcrafted features. Recent studies employ deep learning techniques, particularly convolutional neural networks (CNN), to automatically learn discriminative features from satellite images. For example, some studies \cite{chen2022multi,fan2022multilevel} classify urban villages by constructing various deep learning models over satellite images and street images. Another study \cite{ fan2022urban} classifies urban informal settlements using very high-resolution remote sensing images and time-series population density data. Also a recent work \cite{xiao2023contextual} uses an urban region graph and designs a contextual master-slave framework to effectively detect the urban village. However, these studies focus on image classification and fail to identify urban village boundaries, providing limited information for sustainable cities.

On the other hand, some studies formulate urban village identification as a segmentation problem. For example, the Mask R-CNN model is used to detect urban villages and segment the boundaries of urban villages from satellite images \cite{chen2019identifying}. Another two studies \cite{mast2020mapping,pan2020deep} respectively utilize the well-established semantic segmentation models, including Fully Convolutional Neural Networks (FCN) and U-Net, to map urban village areas in Shenzhen and Guangzhou. Moreover, UVLens \cite{chen2021uvlens} employs taxi trajectories to divide the city-wide satellite image into smaller patches and then incorporates bike-sharing drop-off data into these image patches and utilizes the Mask R-CNN model \cite{he2017mask} to detect urban villages. Overall, existing studies on urban villages either focus narrowly on classification or struggle with inaccurate semantic segmentation. Besides, such studies often rely on additional data sources such as street views and traffic data, which do not apply to all cities.

\subsubsection{SAM Applications.}
Since the proposal in April 2023, SAM has been widely used in different fields, such as medical image processing \cite{ma2023segment,zhou2023can}, 3D vision \cite{cen2023segment,shen2023anything}, inpainting \cite{yu2023inpaint}, object tracking \cite{yang2023track,rajivc2023segment} and so on, which fall into two application ways: (i) Fine-tuning or adding an adapter on SAM image encoder. For example, SAMed \cite{zhang2023customized}, MedSAM \cite{wu2023medical} and 3DSAMadpter \cite{gong20233dsam} entail the customization of SAM specifically for medical image segmentation with adapters incorporated, yielding performance improvement in medical image segmentation tasks. (ii) Generating task-specific prompts. For example, AutoSAM \cite{shaharabany2023autosam} designs an auxiliary convolution network that replaces the prompt embedding for medical imaging domains. RSPrompter \cite{chen2023rsprompter} develops anchor-based and query-based prompts with SAM for satellite image-based instance segmentation. Motivated by such SAM-based applications, we adapt SAM to the urban village identification problem.

\section{Preliminary}
In this section, we provide problem statement and important models of SegFormer and SAM used in the methodology. 

\subsubsection{Problem Statement.}
\newtheorem{problem}{Problem}
Urban village identification refers to the task of identifying and delineating the boundaries of urban villages within a given geographical area, separating them from the surrounding areas. Thus, the research problem with satellite images is formally defined as:
\begin{problem}
Given any satellite image $\mathcal{I}$, the satellite image-based urban village identification problem is to design method $f$ to identify specific boundaries $U$ for urban village therein (if existed), denoted as $U$:$f(I)\rightarrow{U}$.
\end{problem}

\subsubsection{SegFormer.} SegFormer \cite{xie2021segformer} builds an encoder-decoder framework to achieve impressive performance in semantic segmentation tasks. In the encoder part, SegFormer employs a hierarchical pyramid Vision Transformer (ViT) \cite{dosovitskiy2020image} to break down the input image into hierarchical regions and process them at different levels of abstraction. In the decoder part, a multi-layer perceptron  (MLP) is developed to gather information from various layers, effectively merging local attention and global attention mechanisms to create potent representations, which are finally upsampled to produce the ultimate segmentation mask.

\subsubsection{SAM.} SAM \cite{kirillov2023segment} designs a flexible prompting-enabled model architecture for category-agnostic segmentation. To be specific, SAM consists of an image encoder, a prompt encoder, and a mask decoder, where the image encoder is pre-trained using the masked autoencoder technique, the prompt encoder handles dense and sparse inputs, and the mask decoder predicts the masks based on the encoded embeddings. Especially, SAM supports external prompts like boxes, points, and texts for segmenting objects.

\section{Methodology}

\begin{figure}[t]
\centering
\includegraphics[width=0.45\textwidth]{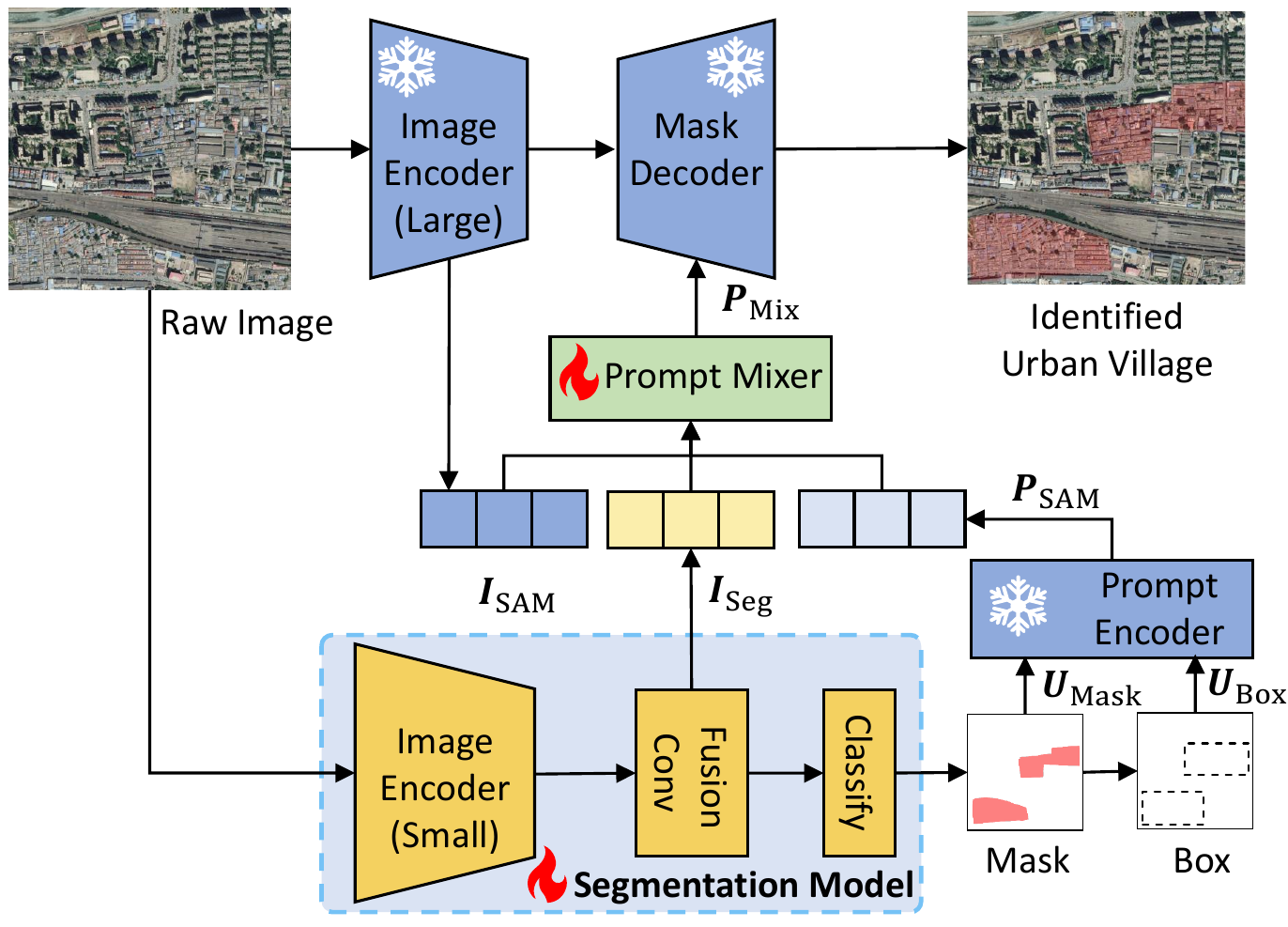} 
\caption{The illustration of proposed UV-SAM framework. The snowflake and torch symbols in the figure signify that the model parameters in this part are kept frozen and learnable, respectively.}
\label{fig:framework}
\end{figure}

\subsubsection{Generalist-Specialist Framework Overview.}
Figure~\ref{fig:framework} presents the main framework of our proposed UV-SAM model for the urban village identification problem, which falls into a generalist-specialist style. Considering the limitations of existing models for urban village identification, which struggle to accurately define the boundaries of urban villages, in the generalist part, we leverage SAM's robust edge detection capabilities to learn such finer boundaries. Moreover, in the specialist part, to provide urban village-specific prompts for SAM, we employ a lightweight semantic segmentation model, SegFormer, for prompt generation.

\subsubsection{Image Encoder.} Specifically, given a satellite image $I$ as input, the image is fed into SAM's image encoder $\Phi_{\text{SAM-Image}}$ with large pre-trained parameters and SegFormer's image encoder $\Phi_{\text{Seg-Image}}$ with small learnable parameters, with output as $\bm{I}_{\text{SAM}}$ and $\bm{I}_{\text{Seg}}$, respectively. Since $\bm{I}_{\text{Seg}}$ comprises features of multiple scales, UV-SAM applies a fusion layer of MLP $\Phi_{\text{Agg}}$ to aggregate such features. The above processes are expressed as:
\begin{align}
    &\bm{I}_{\text{SAM}}=\Phi_{\text{SAM-Image}}(I)\\
    &\bm{I}_{\text{Seg}}=\Phi_{\text{Agg}}(\Phi_{\text{Seg-Image}}(I))
\end{align}

Subsequently, the $\bm{I}_{\text{Seg}}$ goes through the classification layer within SegFormer to generate masks $U_{\text{Mask}}$ for urban villages therein. After undergoing image processing, the masks are then used to derive the corresponding bounding boxes $U_{\text{Box}}$. This process can be expressed as:
\begin{align}
    &U_{\text{Mask}}=\text{MLP}(\bm{I}_{\text{Seg}})\\
    &U_{\text{Box}}\underset{\text{Image Process}}{\longleftarrow} U_{\text{Mask}}
\end{align}

\subsubsection{Prompt Generation.} The above masks and bounding boxes are utilized as prompts, and fed into the prompt encoder $\Phi_{\text{SAM-Prompt}}$ resulting in a sparse prompt embedding $\bm{P}_{\text{SAM}}$, which encompasses explicit prompt information on urban village location details, which is expressed as:
\begin{align}
 \bm{P}_{\text{SAM}}=\Phi_{\text{SAM-Prompt}}(U_{\text{Box}}, U_{\text{Mask}})
\end{align}

Moreover, since both $\bm{I}_{\text{SAM}}$ and $\bm{I}_{\text{Seg}}$ aggregate abstract semantic information specific to urban villages, UV-SAM also models both as semantic prompts. Especially, a prompt generation module is designed to fuse such four types of prompts together, whose two prompt mixing variants of $\Phi^\text{Add}_{\text{Mix}}$ and $\Phi^\text{MLP}_{\text{Mix}}$ can be expressed as:
\begin{align}
& \bm{P}_{\text{Mix}}\!\!=\!\!\Phi^{\text{Add}}_{\text{Mix}}(\bm{P}_{\text{SAM}},\bm{I}_{\text{Seg}},\bm{I}_{\text{SAM}})\!\!=\!\!\bm{P}_{\text{SAM}}+\bm{I}_{\text{Seg}}+\bm{I}_{\text{SAM}}\\
& \bm{P}_{\text{Mix}}\!\!=\!\!\Phi^{\text{MLP}}_{\text{Mix}}(\bm{P}_{\text{SAM}},\bm{I}_{\text{Seg}},\bm{I}_{\text{SAM}})\!\!=\!\!\text{MLP}([\bm{P}_{\text{SAM}};\bm{I}_{\text{Seg}};\bm{I}_{\text{SAM}}])
\end{align}
where $\Phi^\text{Add}_{\text{Mix}}$ involves directly adding normalized features and ensures a straightforward fusion of insights, while $\Phi^\text{MLP}_{\text{Mix}}$ entails concatenating normalized features and then passing them through a projection head for dimensionality reduction, offering a more intricate yet controlled method of knowledge synthesis. The choice between the two forms could depend on the nature of the information being merged and the specific requirements of the task. 

\subsubsection{Mask Decoder.} Finally, based on the mixed prompt and pre-trained mask decoder in SAM, UV-SAM identifies urban villages in satellite images as follows:
\begin{align}
    U=\Phi_{\text{SAM-Mask}}(\bm{I}_{\text{SAM}},\bm{P}_{\text{Mix}})
\end{align}
where elements in $U$ identify whether specific pixels belong to urban villages.   

\subsubsection{Training Loss.}
Similar to SAM, in the context of the larger model, we adopt a mask prediction strategy involving a linear combination of focal loss $\mathcal{L}_{\rm focal}$ \cite{lin2017focal}, dice loss $\mathcal{L}_{\rm dice}$ \cite{milletari2016v} and mean-square-error loss $\mathcal{L}_{\rm mse}$ at a weight of 1:1:1. In addition, SegFormer continues to utilize the straightforward cross-entropy loss $\mathcal{L}_{\rm Seg}$ for its loss function. Consequently, the overall loss can be expressed as follows:
\begin{align}
 \mathcal{L}_{\rm SAM}&=\mathcal{L}_{\rm focal}+\mathcal{L}_{\rm dice}+\mathcal{L}_{\rm mse}\\
 \mathcal{L}&=\lambda \mathcal{L}_{\rm SAM}+\mathcal{L}_{\rm Seg}
\end{align}
where $\lambda$ is a hyper-parameter to weigh the impacts of generalist and specialist modules.

\section{Experiments}
In this section, we conduct experiments to answer the following research questions:
\begin{itemize}
    \item \textbf{RQ1:} How does our proposed UV-SAM model perform compared with existing baseline approaches?
    \item \textbf{RQ2:} What is the effectiveness of each designed module in our proposed UV-SAM model? 
    \item \textbf{RQ3:} Can our proposed UV-SAM model identify the spatial distribution of urban villages?
    \item \textbf{RQ4:} Can our proposed UV-SAM model identify the evolving trends of urban villages with area and amount?
\end{itemize}

\subsection{Experiment Setups}
\subsubsection{Datasets.}
For our research, we collected datasets consisting of satellite images from two major cities of Beijing and Xi'an in China. Table~\ref{tab:data} reports basic statistics for datasets. For training purposes, we randomly split the dataset into three subsets of training, validation, and testing sets with the proportion of 6:2:2.

\begin{table}[htbp]
\centering
\begin{tabular}{cccc}
\toprule
City & \#Satellite Image &\#Urban Village & Year  \\
\midrule
Beijing  & 2,491 & 545 &2016\\
Xi'an  & 837 & 205 &2018\\

\bottomrule
\end{tabular}
\caption{Dataset statistics.}
\label{tab:data}
\end{table}

We specifically focus on main urban areas to capture the dynamics of urban village evolution. The satellite images are obtained from ArcGIS\footnote{\url{https://geoenrich.arcgis.com/}} and have a resolution of approximately 1.05 meters per pixel. To prepare the dataset for training and evaluation, we merge the individual satellite images into larger images of 1024 $\times$ 1024 pixels. This merging process ensures that the resulting images contained comprehensive main urban area information. 

As for labels, to begin with, we recruited a group of participants from urban research and provided them with appropriate incentives. We also conducted training sessions to ensure that the participants had a good understanding of urban villages and related background knowledge. To facilitate the labeling process, we utilized the EasyData\footnote{\url{https://ai.baidu.com/easydata/}} crowdsourcing platform, through which we randomly assigned image patches to the participants for cross-validation. Each patch was assigned to three participants to ensure accuracy and consistency in the labeling process. To maintain quality control, we assigned specific individuals to validate the mask annotations provided by the participants. This validation process helped to ensure the accuracy and reliability of the obtained mask labels. By conducting per-pixel voting, we obtained the ground truth labels for all the image patches.

\subsubsection{Baselines.}
We have conducted a comparative analysis of our model against various baseline approaches:
\begin{itemize}
    \item \textbf{FCN} \cite{long2015fully}. FCN replaces fully connected layers with convolutional layers, enabling end-to-end pixel-wise predictions.
    \item \textbf{DeepLabv3+} \cite{chen2018encoder}. The architecture of DeepLabv3+ is enhanced by integrating atrous spatial pyramid pooling and decoder modules, resulting in a more sophisticated design.
    \item \textbf{UVLens} \cite{chen2021uvlens}. UVLens integrates bike-sharing drop-off data and satellite images into image patches and applies Mask R-CNN \cite{he2017mask} model for urban village identification.
    \item \textbf{RSPrompter} \cite{chen2023rsprompter}. RSPrompter incorporates elements from both Faster R-CNN \cite{he2016deep} and Transformer \cite{vaswani2017attention} architectures into the prompt generation process for satellite image instance segmentation.
\end{itemize}

\subsubsection{Metrics.}
\begin{table*}
\centering
\begin{tabular}{c|cccc|ccccc}
\toprule
\textbf{Dataset} & \multicolumn{4}{c|}{\textbf{Beijing}} &\multicolumn{4}{c}{\textbf{Xi'an}} \\
\midrule
Method  &IoU&F1-Score&Recall&Precision  &IoU&F1-Score&Recall&Precision  \\
\midrule
FCN & \underline{0.660} &\underline{0.802} &0.752 &\underline{0.859}& \underline{0.720} &0.833 &0.800 &\underline{0.870}  \\
DeepLabv3+ &  0.650 & 0.787&0.719& \textbf{0.870}& 0.668 & 0.821&0.780& 0.867\\
UVLens & 0.623 &0.783 & 0.777&0.790& 0.687 &\underline{0.863} & \underline{0.880}&0.867\\ 
RSPrompter & 0.462 &0.687 &\underline{0.860} &0.571 & 0.568 & 0.800& 0.800 & 0.800 \\
\midrule
UV-SAM & \textbf{0.721}& \textbf{0.871} &\textbf{0.893}& 0.851 & \textbf{0.747}& \textbf{0.904} &\textbf{0.940}& \textbf{0.871}\\
\bottomrule
\end{tabular} 
\caption{Overall performance of UV-SAM and baselines on two datasets. Bold denotes the best results and underline denotes the second-best results.}%
\label{table:beijing_xian}
\end{table*}

To evaluate the accuracy of our identification method, we compare the segmented urban villages with the ground truth dataset under two types of metrics in respect to detection accuracy and segmentation accuracy.

For detection accuracy, if a detected urban village spatially overlaps with an urban village in the ground-truth dataset, we mark it as true positive, otherwise false positive, based on which we calculate precision, recall and F1-score.

For segmentation accuracy, we utilize the widely-used Intersection over Union (IoU) metric, which is calculated as the intersection area divided by the union area between the segmented urban villages and the corresponding ground-truth urban villages. 

\subsubsection{Implementation.}
In our experiments, we consistently employ the ViT-Large backbone of SAM and the MiT-B0 lightweight encoder of SegFormer, unless specified otherwise.
We select the Adam optimizer to facilitate parameter learning and incorporate a cosine annealing scheduler to gradually decrease the learning rate. The mini-batch size is fixed at 4, and the complete training process spans 100 epochs. We conduct a grid search for optimal values of the learning rate, weight decay, and $\lambda$, from \{0.005, 0.0005, 0.00005\}, \{0.01, 0.001\} and \{0.1, 1, 10\}, respectively. Besides, based on the validation performance, we select $\Phi^{\text{MLP}}_{\text{Mix}}$ and $\Phi^{\text{Add}}_{\text{Mix}}$ for Beijing and Xi'an, respectively. The experiment details are available at the link\footnote{\url{https://github.com/tsinghua-fib-lab/UV-SAM}}.

\begin{table*}
\centering
\begin{tabular}{c|cccc|ccccc}
\toprule
\textbf{Dataset} & \multicolumn{4}{c|}{\textbf{Beijing}} &\multicolumn{4}{c}{\textbf{Xi'an}} \\
\midrule
Variants  &IoU&F1-Score&Recall&Precision  &IoU&F1-Score&Recall&Precision  \\
\midrule
w/o Box & 0.708 &0.838 &0.876 & 0.803& 0.173&0.267 &0.160 &0.800  \\
w/o Mask &  0.635 & 0.842&0.860& 0.825& 0.744 & 0.876&0.920& 0.836\\
w/o SAM emb  & 0.697 &0.846 &0.909 &0.791 & 0.717 &0.826 &0.900 &0.763\\

w/o Seg emb & 0.694 &0.832 &0.901 &0.773 & 0.733 & 0.849& 0.900 & 0.804 \\
w/o SAM  & 0.688 &0.854 &0.893 &0.818 & 0.731 &0.860 &0.860 &0.860\\
\midrule
UV-SAM & \textbf{0.721}& \textbf{0.871} &\textbf{0.893}& \textbf{0.851} & \textbf{0.747}& \textbf{0.904} &\textbf{0.940}& \textbf{0.871}\\
\bottomrule
\end{tabular} 
\caption{Ablation study of UV-SAM variants on two datasets.}
\label{table: Ablation study}
\end{table*}

\subsection{Overall Performance (RQ1)}
Table~\ref{table:beijing_xian} shows the overall performance comparison on Beijing and Xi'an datasets. From these results, we have the following observations:
\begin{itemize}
    \item \textbf{UV-SAM achieves the best performance across both datasets.}~The results showcase that our proposed model achieves state-of-the-art performance, which successfully adapts SAM into urban village identification. For the segmentation accuracy, compared with the baselines, our model outperforms the best baseline by 4\%-9\% on IoU in two datasets. Similar performance improvements can be also found in F1-score for the detection accuracy. It is notable that performance difference on two datasets with DeepLabv3+. This is due to how well DeepLabv3+’s structure matches the characteristics of the Beijing dataset. The DeepLabv3+ architecture combines high-level features for semantic information with low-level features for capturing boundary details. Beijing’s unique features, with denser traditional courtyard style housing and shorter buildings, differ from the high-rise, dense buildings of Xi’an. In addition, owing to the generalized generalist-specialist framework, all baselines in Table~\ref{table:beijing_xian} can be incorporated into UV-SAM as the specialist module, which can bring further performance improvement for urban village identification.
    
    \item \textbf{Existing SAM-based models perform poorly in urban village identification.} According to the results in Table~\ref{table:beijing_xian}, the performance of RSPrompter \cite{chen2023rsprompter} notably lags behind that of other baseline models in terms of IoU and F1-score, e.g., the worst IoU and F1-score of 0.462 and 0.687 in Bejing dataset. Such results suggest that the learnable prompts in RSPrompter fail to capture the intricate and abstract semantic features that hold particular relevance to urban villages, and thus provide useless guidance to SAM. Besides, the performance drop also emphasizes the non-tricky adaption of SAM to urban village identification. 

    \item \textbf{Transformer-based encoders demonstrate a better semantic understanding of urban villages.} In the capacity of transformer-based architectures, our proposed UV-SAM exhibits remarkable superiority compared with other CNN-based models, in terms of IoU and F1-score metrics. As described before, urban villages embody intricate and advanced semantic concepts, and the demarcation of their boundaries within satellite imagery is notably influenced by contextual factors in their surrounding environment. Thus, the Transformer architecture with the attention mechanism can better capture fine-grained features therein, while CNN-based models mainly grasp higher-level semantic abstractions, leading to inaccurate boundaries and inferior performance.
\end{itemize}

\begin{figure}[t]
    \centering
    \subfigure[Beijing]{
\includegraphics[width=0.226\textwidth]{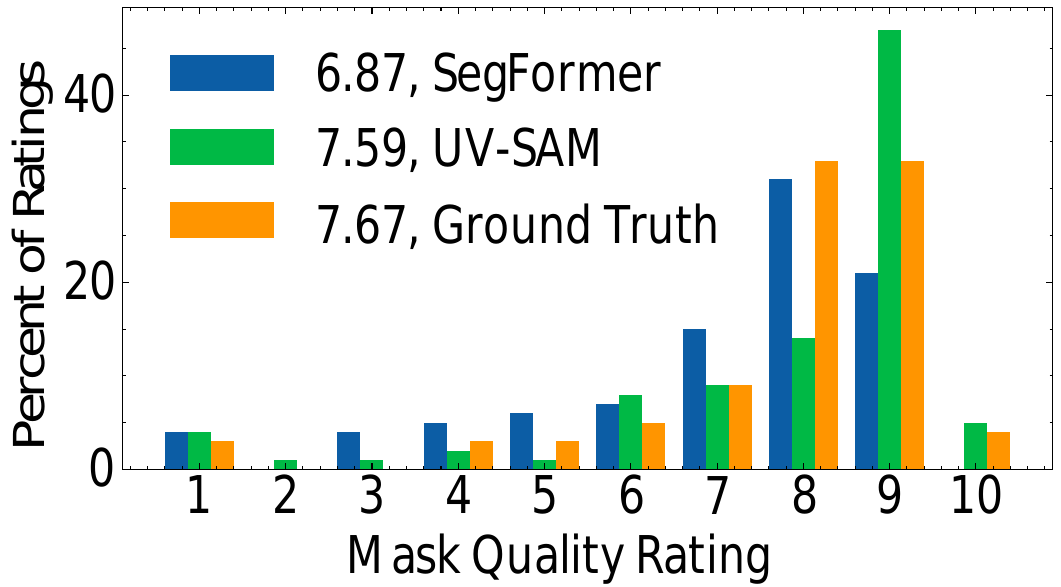}
    }
    \subfigure[Xi'an]{
\includegraphics[width=0.226\textwidth]{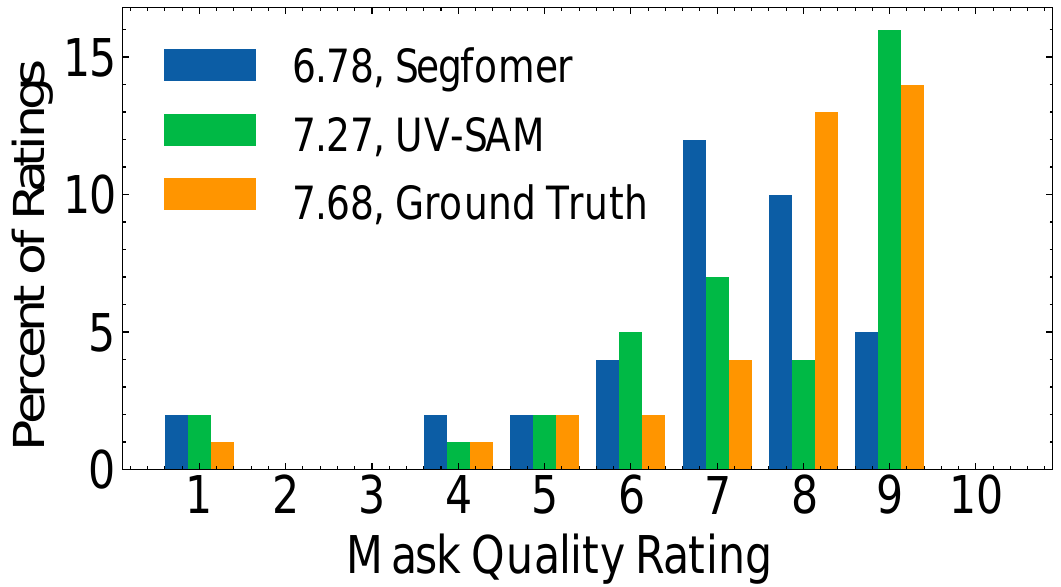}}
    \caption{Mask quality rating distributions by datasets from our human evaluation study in Beijing and Xi'an, with average scores shown in the legend.}
    \label{fig:beijing_mask_quaility}
\end{figure}
\vspace{0.12cm}
Furthermore, to better evaluate the mask quality in urban villages, we introduce the human study for evaluation \cite{kirillov2023segment}. Specifically, we present masks generated by models to annotators and require them to rate the quality of each mask from 1 to 10. A score of 1 means that the mask has no relevance to urban villages while 10 indicates that there are no noticeable errors in the identified boundaries of urban village areas. We conduct a comparison between the masks produced by SAM and those generated by SegFormer, along with the ground truth data, which are presented in Figure~\ref{fig:beijing_mask_quaility}. The results show that \textbf{UV-SAM achieves better mask quality than SegFormer across both datasets}. For example, on the Beijing dataset, UV-SAM achieves an average rating of 7.59, while SegFormer is 6.87, compared with the ground truth of 7.67. Within the lower score range, UV-SAM's performance falls short of the baseline counterpart. Conversely, in the higher score range, there is a conspicuous increase in frequency. Such results demonstrate the effectiveness of SAM for segmenting boundaries.
\vspace{-0.55cm}
\subsection{Ablation Study (RQ2)}
\vspace{-0.53cm}
To evaluate the effectiveness of each module in UV-SAM, Table \ref{table: Ablation study} shows the detection and segmentation performance of different model variants on both datasets.
According to the results, without the box prompt, our model performance drops 1.8\% and 75.9\%. Thus, the box prompt plays an important role in the performance guarantee, which guides the mask decoder of the SAM to focus on the regions of interest. Besides, the performance drop on Xi'an dataset can be largely attributed to the straightforward way of prompt addition. Moreover, the mask prompt offers a dense embedding that specifically emphasizes the boundaries of objects within the image, contributing 11.9\% and 0.4\% on IoU for two datasets, respectively. Furthermore, with abstract semantic information specific to urban villages, the SAM embedding from the image encoder (large) further achieves 3\%-4\% improvement on IoU. Finally, our model can get an improvement of 2\%-3\% with the assistance of segmentation embedding obtained from the image encoder(small), capturing the high-level semantics provided by the specialist-like semantic segmentation model. Thus, all four types of prompts are essential for effective urban village identification. In addition, our model performance drops 4.6\% and 2.1\% without the SAM. So a generalist-like SAM provides more accurate boundary information for urban village identification.

\begin{figure}[htbp]
\centering
\includegraphics[width=0.40\textwidth]{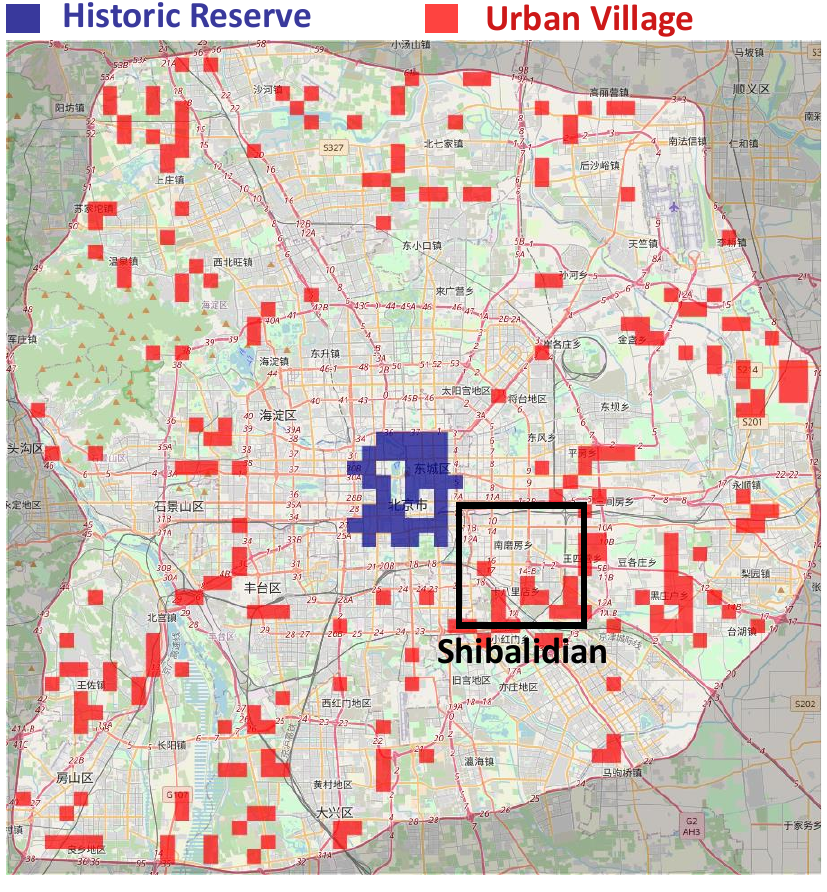}\caption{ Urban village (UV) distribution in Beijing in 2020.}
\label{fig:beijing_spatial}
\end{figure}

\subsection{Spatial Distribution Analysis (RQ3)}
To mitigate the potential risks of urban villages to urban development as well as improve citizens' living conditions, the governments often initiate gradual demolitions and resident relocations therein. Thus, identifying the spatial distribution of urban villages offers a crucial reference point for urban planning. In Figure~\ref{fig:beijing_spatial}, we visualize the spatial distribution of urban villages within the sixth ring road of Beijing in 2020. 

As depicted in the figure, within the second ring road of Beijing, numerous historical reserve blocks are presented, which often consist of courtyard-style housing, accommodating a few households in close proximity. Despite their historical and conservation value, these areas typically have small per capita living spaces, poor sanitation conditions, and low greenery coverage, which align well with the definition of urban villages. On the contrary, urban villages are found to be distributed more thinly between the third and sixth ring roads. Particularly, there is a noticeable clustering of urban villages near Shibalidian Township\footnote{\url{https://en.wikipedia.org/wiki/Shibalidian}}, which is a famous urban village cluster in Beijing. Moreover, the southern and eastern parts of this area show a higher density of urban villages compared to the western and northern parts. This discrepancy in distribution could be attributed to historical population movement patterns and local levels of economic development.

\begin{figure}[t]
\centering
\includegraphics[width=0.4\textwidth]{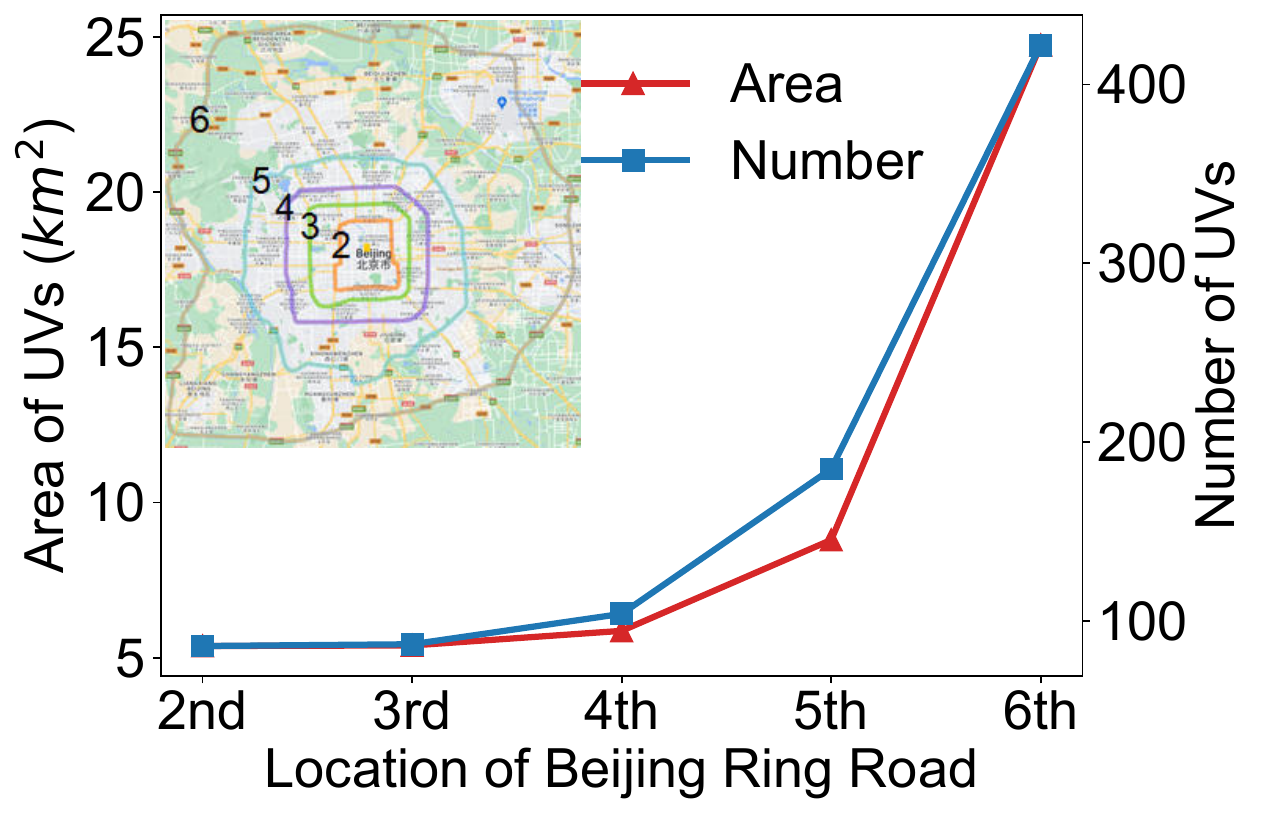}\caption{Urban village (UV) distribution along Beijing's ring roads with respect to area and amount in 2020.}
\label{fig:beijing_spatial_trend}
\end{figure}

To quantify the spatial distribution, we further plot the urban village distribution curves along Beijing's ring roads with respect to area and amount in Figure~\ref{fig:beijing_spatial_trend}. We crudely determined the count and extent of urban villages by utilizing the number of predicted masks and their cumulative pixel values derived from satellite imagery results. According to the results, there is a significant increase in both area and number of urban villages between the fifth and sixth ring roads, where the distance to the urban center is far enough and the buildings from the original village are preserved.

\begin{figure}[htbp]
    \centering
    \subfigure[Beijing]{
        \includegraphics[width=0.22\textwidth]{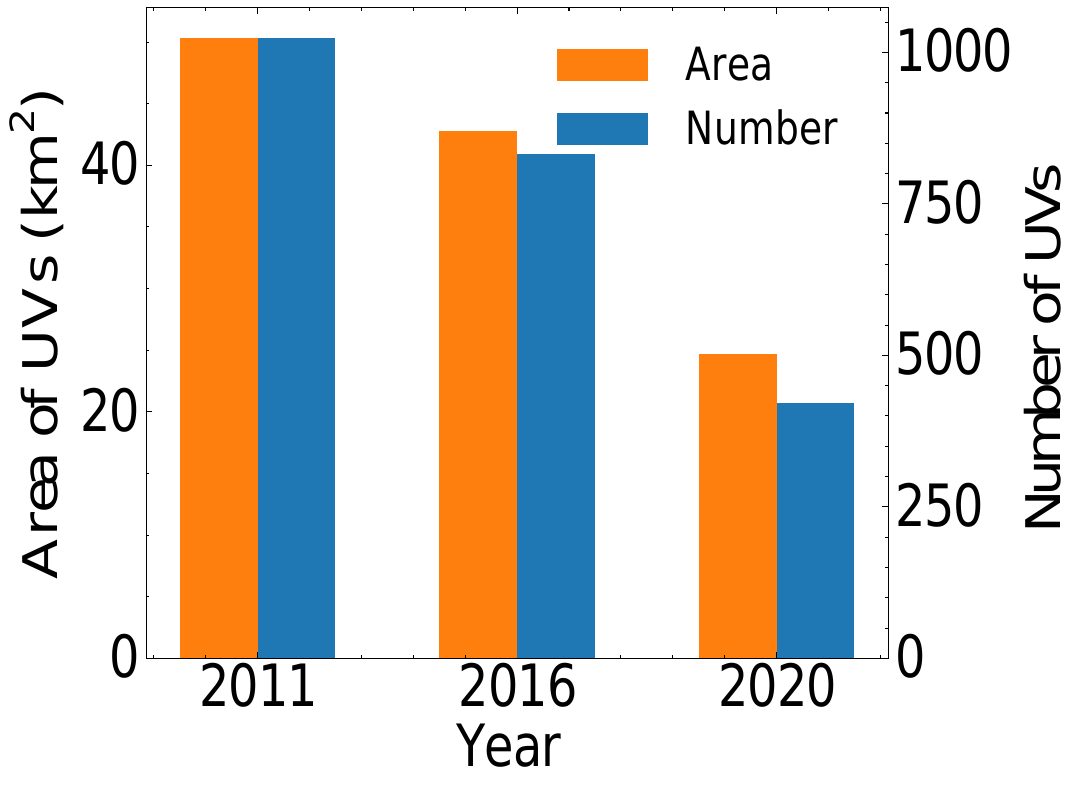}
    }
    \subfigure[Xi'an]{
    \includegraphics[width=0.22\textwidth]{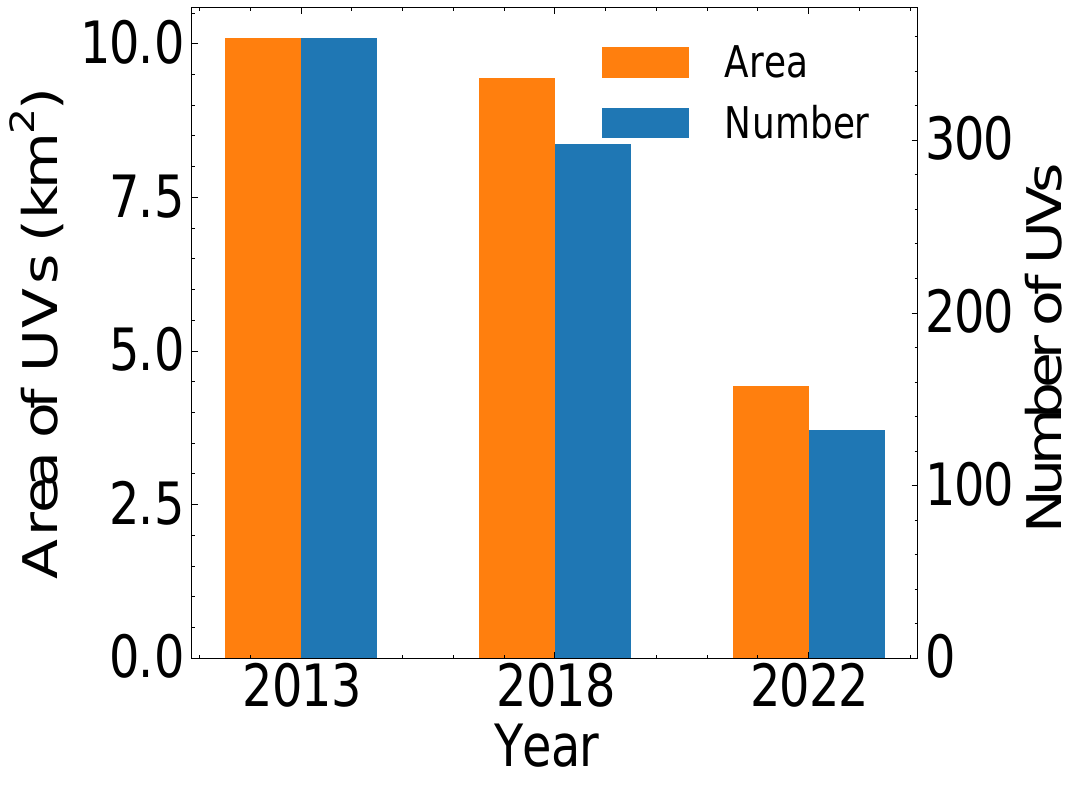}}
    \caption{The change of the area and number of urban villages over years in Beijing and Xi'an. }
    \label{fig:beijing_xian_area_num}
\end{figure}

\subsection{Evolving Trend (RQ4)}
\begin{figure}[t]
    \centering
    \subfigure[The Jijiamiao Village in 2011]{
        \includegraphics[width=0.3\linewidth]{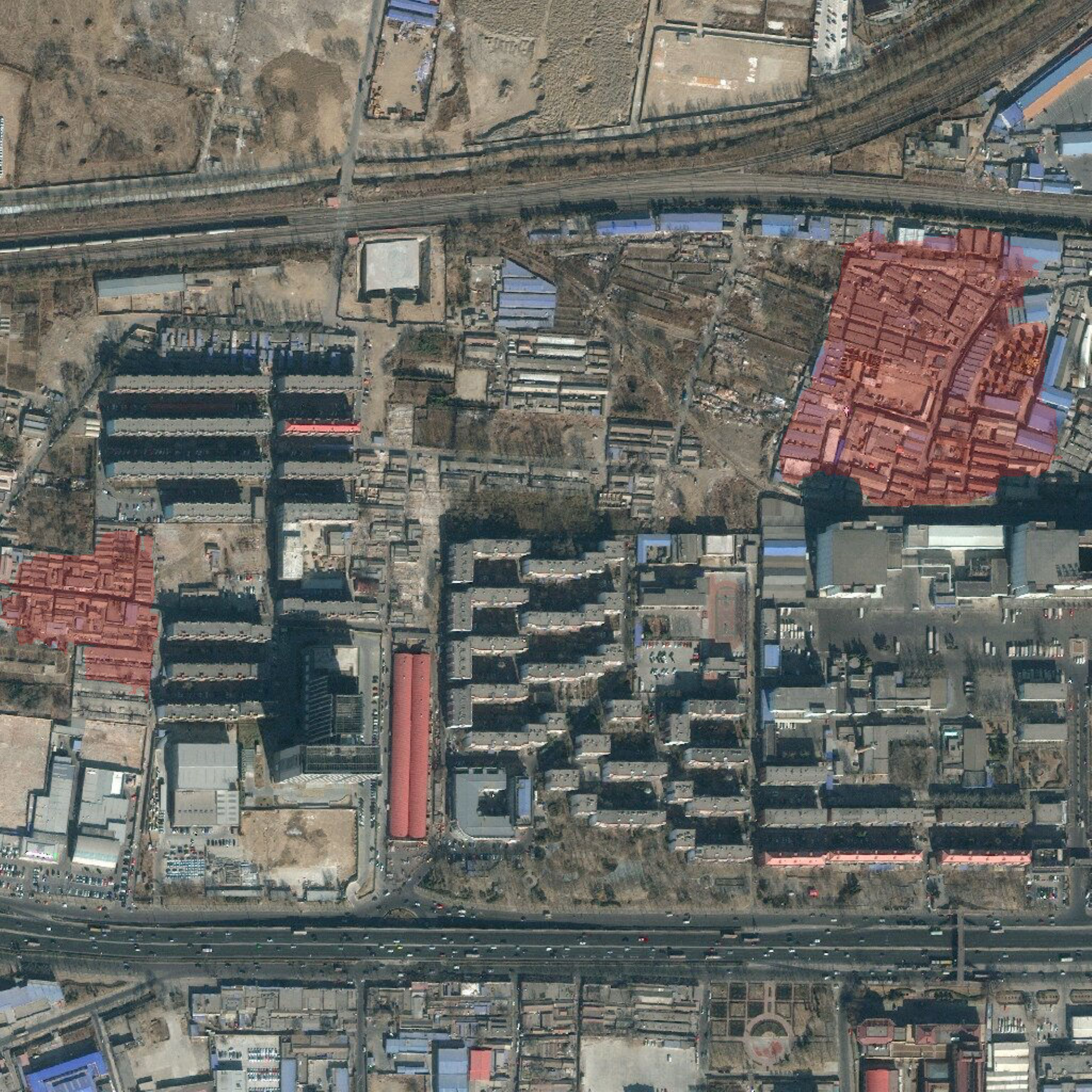}
    }
    \subfigure[The Jijiamiao Village in 2016]{
    \includegraphics[width=0.3\linewidth]{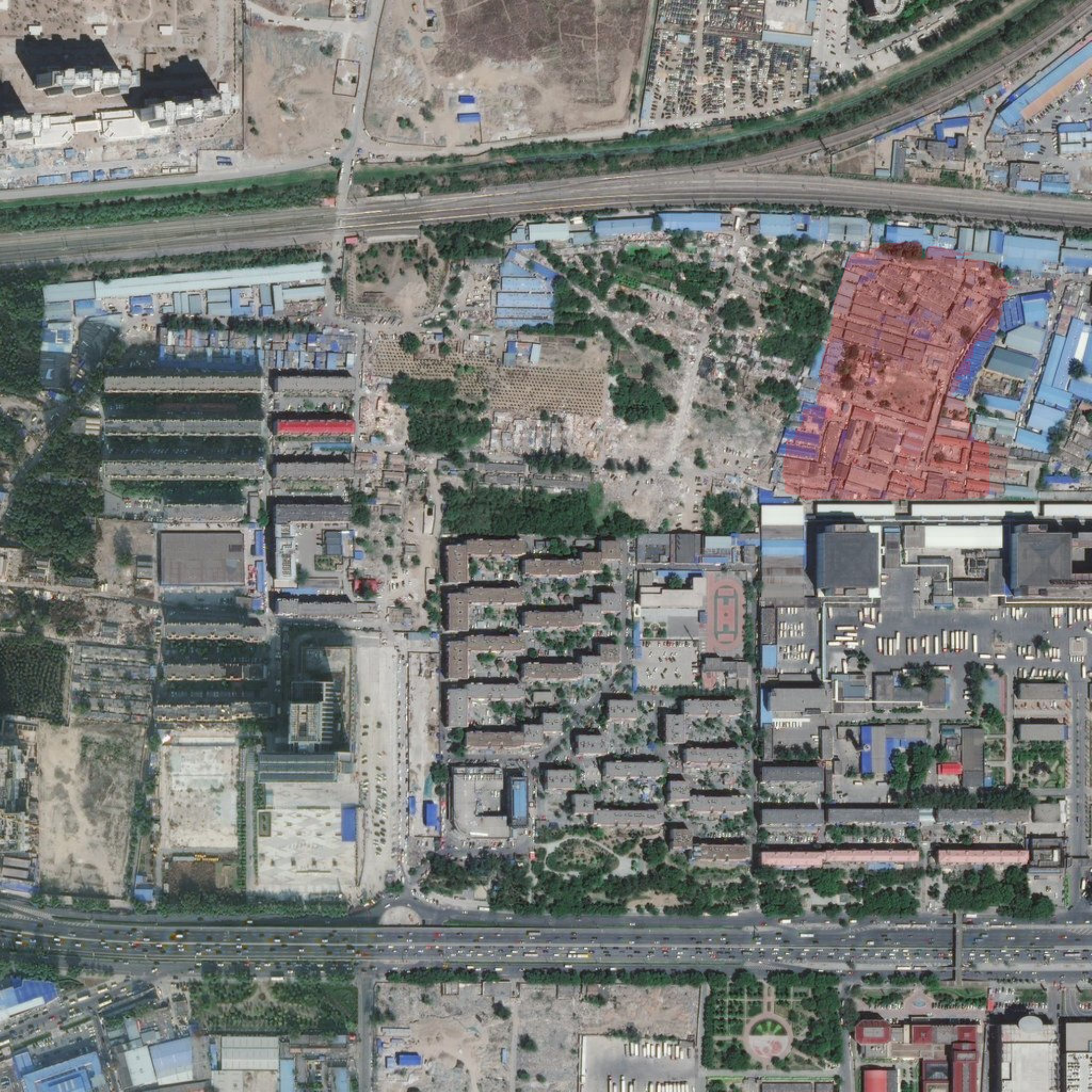}
    }
    \subfigure[The Jijiamiao Village in 2020]{
    \includegraphics[width=0.3\linewidth]{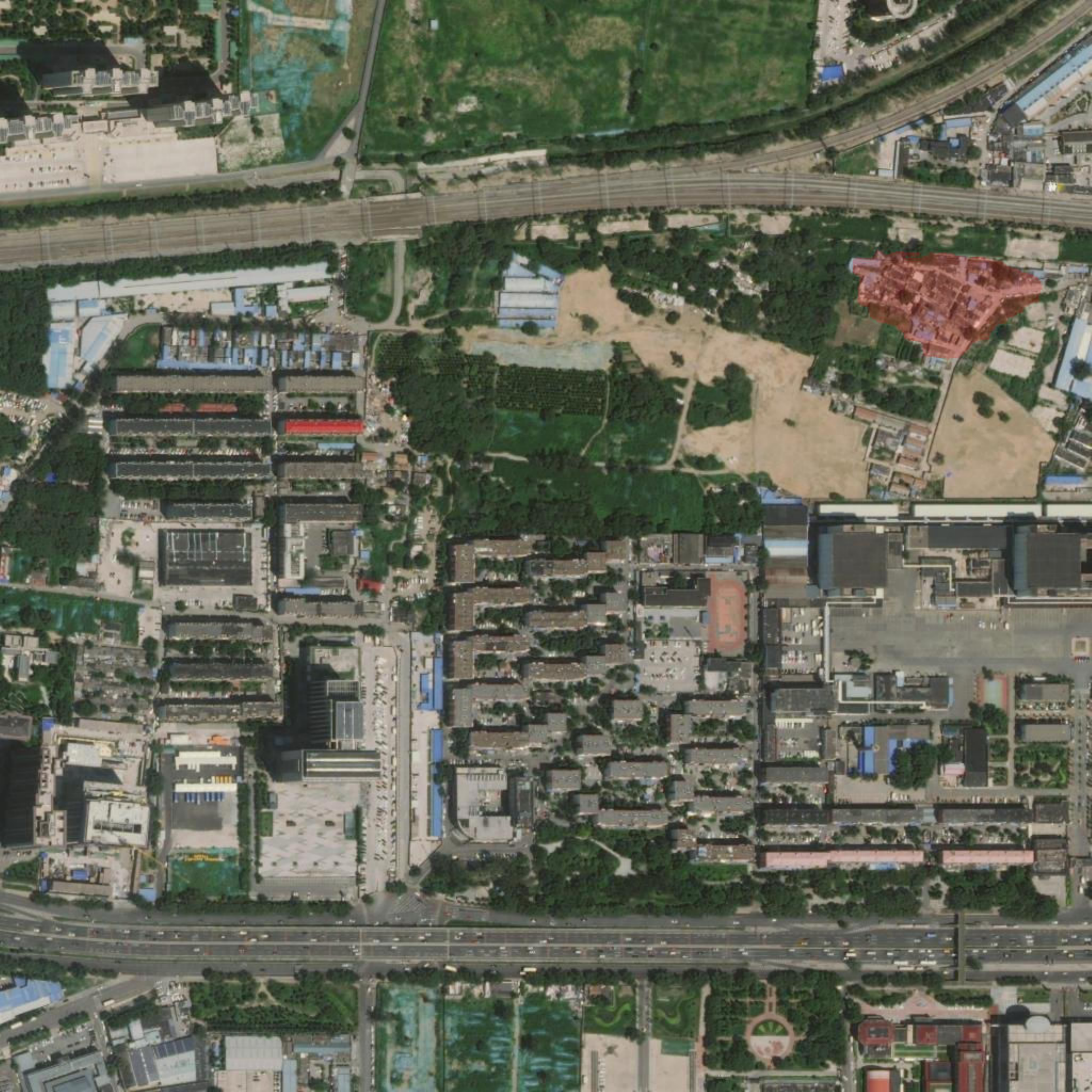}
    }
    \caption{The boundary change of the Jijiamiao Village from 2011 to 2020. The highlighted red regions denote the areas where urban villages have been identified, signifying a gradual reduction over time.}
    \label{fig:beijing_change}
\end{figure}

To understand the formation, expansion and shrinkage of urban villages, in Figure~\ref{fig:beijing_change}, we illustrate the variations in urban village area and quantity between different years for the cities of Beijing and Xi'an, using the satellite images captured at various time points.

According to the results, the city of Beijing hosted an estimated 1,000 urban villages in 2011, and Xi'an accommodated around 360 urban villages in the year 2013. By 2016 or 2018, the urban villages only decreased by less than 10 square kilometers. However, a notable transformation occurred by the year 2020. During this time, both the spatial extent and numerical prevalence of urban villages underwent a remarkably rapid contraction, resulting in a reduction of fifty percent compared to their previous levels. This discernible trend is plausibly attributable to the promulgation of the Beijing Urban Master Plan(2016-2035) by the governmental authorities.

Especially, the example of Jijiamiao Village serves as a case currently undergoing transformation\footnote{\url{http://zjw.beijing.gov.cn/bjjs/gcjs/zdgcjs/2016/xmjh/363391/index.shtml}}. As shown in the Figure ~\ref{fig:beijing_change}, the Jijiamiao Village is surrounded by high-rise buildings, their outdated structures no longer in sync with the modern landscape. As early in 2011, policies were introduced to gradually renovate these aging structures. Therefore, by 2016, their presence had diminished compared to 2011. By 2020, they had nearly disappeared entirely. The surrounding green spaces and high-rise buildings are also undergoing slow but steady development.

\section{Conclusion}
In this paper, we propose UV-SAM, a vision foundation model-based framework for urban village identification. The UV-SAM framework introduces a specialist-like semantic segmentation model to generate four types of urban village-specific prompts and then feeds into a generalist-like SAM model to identify urban village boundaries from satellite images. Through comprehensive experiments, we substantiate the effectiveness of our model across various datasets, which also provide deep insights into the spatial distributions and temporal trends of urban villages. Moreover, our study demonstrates the possibility of vision foundation models for sustainable development goals and sustainable cities. 

Despite surpassing baseline performance, it's noteworthy that our results may exhibit a certain degree of reduced interpretability. Thus, in future work, we aim to delve into the intricate interplay of features that underlie the emergence and dissolution of urban villages. We also plan to transfer the proposed framework to slum identification in cities and help understand the global informal settlements.

\section{Acknowledgements}
This work was supported in part by the National Key Research and Development Program of China under 2022YFF0606904, the National Natural Science Foundation of China under U22B2057 and U21B2036.

\newpage
\appendix
\section{EXPERIMENT DETAILS}
\subsection{Experiment Details for Spatial  Distributions and  Evolving Trends}
Here we introduce the details of the pre-classification module for spatial distributions and evolving trends. 
Due to the limited size of the dataset, we introduced a binary classification model specifically to distinguish urban and non-urban areas before the semantic segmentation process. The module is only applied in the spatial analysis of specific cities and is independent of our UV-SAM framework. 

\subsubsection{Datasets.} 
For the training classification model, we construct a dataset for each specific city. The urban village area is used as positive samples and the non-urban village areas are randomly selected as negative samples, ensuring a near 1:1 positive-negative ratio in the training, validation, and test sets.

\subsubsection{Implementation.}
ResNet50 \cite{he2016identity} is used to implement the classification model. We select the Adam optimizer to facilitate parameter learning and incorporate a cosine annealing scheduler to gradually decrease the learning rate. The learning rate is set as 0.0001, and the batch size is fixed at 32. To quantitatively measure the performance of the classification model, we adopt the Area Under Curve (AUC), Recall, Precision, and F1-score as evaluation metrics. 

\begin{table}[H]
\centering
\begin{tabular}{c|cccc}
\toprule
Dataset  &AUC&F1-Score&Precision& Recall  \\
\midrule
Beijing & 0.964 &0.868 & 1.00& 0.767 \\
Xi'an & 0.756 &0.677&0.786&0.595\\
\bottomrule
\end{tabular} 
\caption{Performance of classification module on two datasets.}
\label{table:classify_performance}
\end{table}
\subsubsection{Performance. }
Table~\ref{table:classify_performance} shows the performance of Beijing and Xi'an datasets. We note a considerable level of precision achieved in the Beijing dataset, whereas the Xi'an dataset demonstrates noticeably lower precision. This substantial contrast in results can be attributed to a combination of key factors, notably including the limited scope of the Xi'an dataset and the inherent differences in the quality of available satellite imagery.
\begin{figure}[htbp]
\centering
\includegraphics[width=0.4\textwidth]{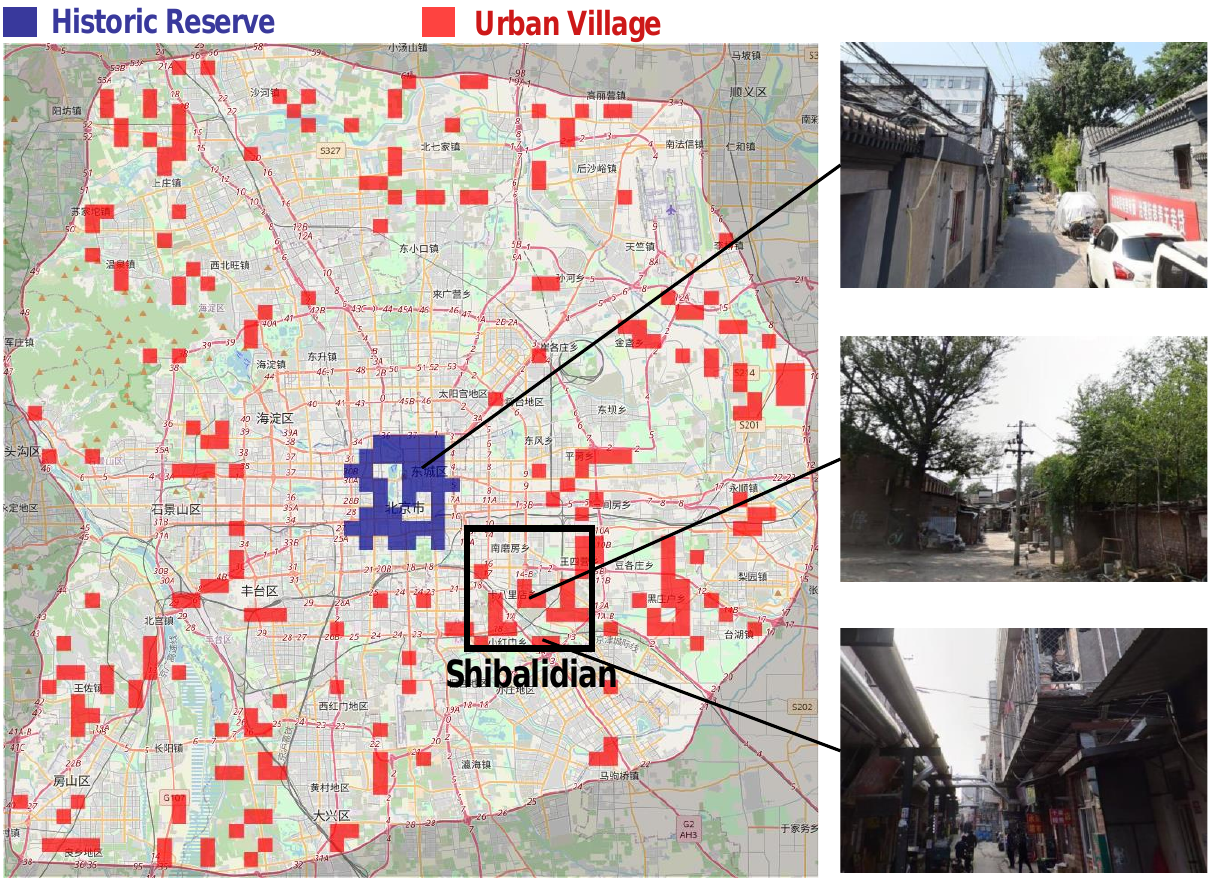}
\caption{ Urban village distribution with street view images in Beijing in 2020.}
\label{fig:beijing_spatial_st}
\end{figure}

\subsection{Spatial Distribution Analysis with Street View Images in Beijing}
Due to the high cost of collecting street view imagery, obtaining street view images for any given time and location is not feasible. Consequently, we can only employ such imagery as an auxiliary tool to facilitate the observation of urban village evolution. 

As illustrated in Figure \ref{fig:beijing_spatial_st}, we choose three street view images from both the historical reserve area and the urban village cluster of Shibalidian, respectively. These images are intended to showcase three distinctly different styles of urban village environments.
\begin{figure}[htbp]
    \centering
    \subfigure[The Jijiamiao Village in 2013]{
        \includegraphics[width=0.3\linewidth]{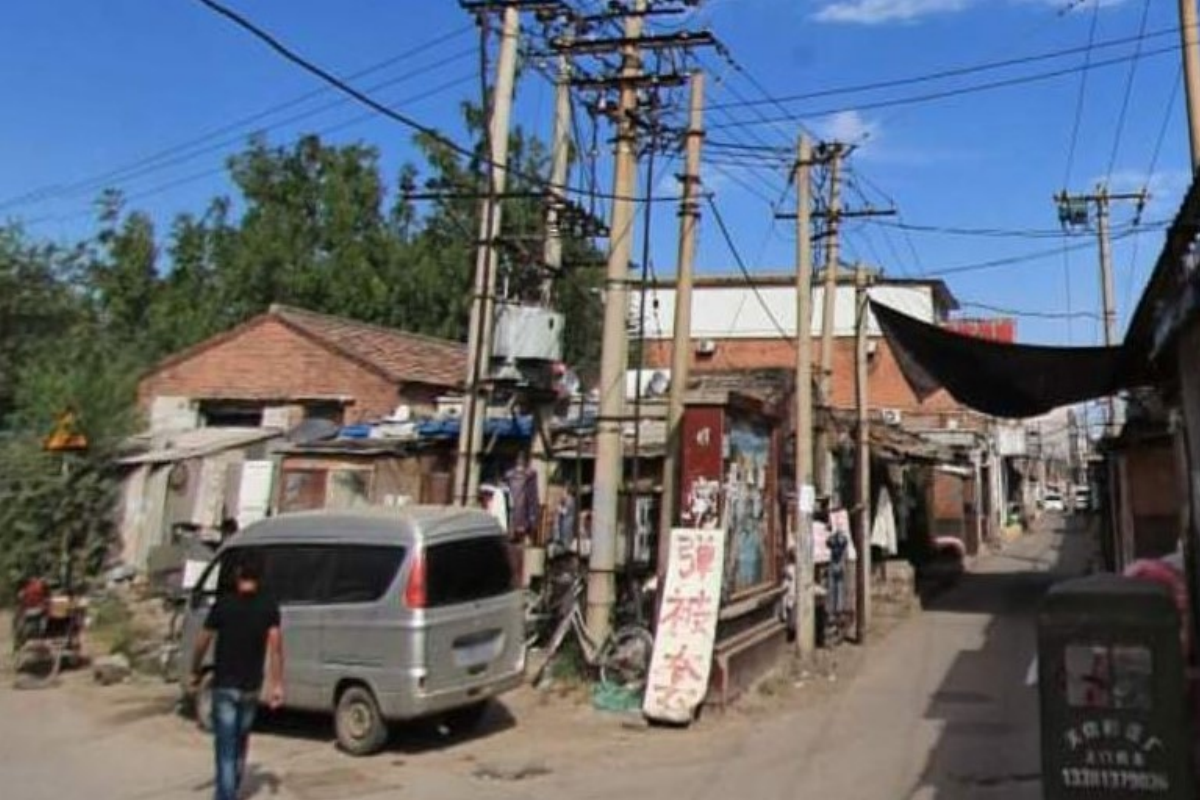}
    }
    \subfigure[The Jijiamiao Village in 2015]{
    \includegraphics[width=0.3\linewidth]{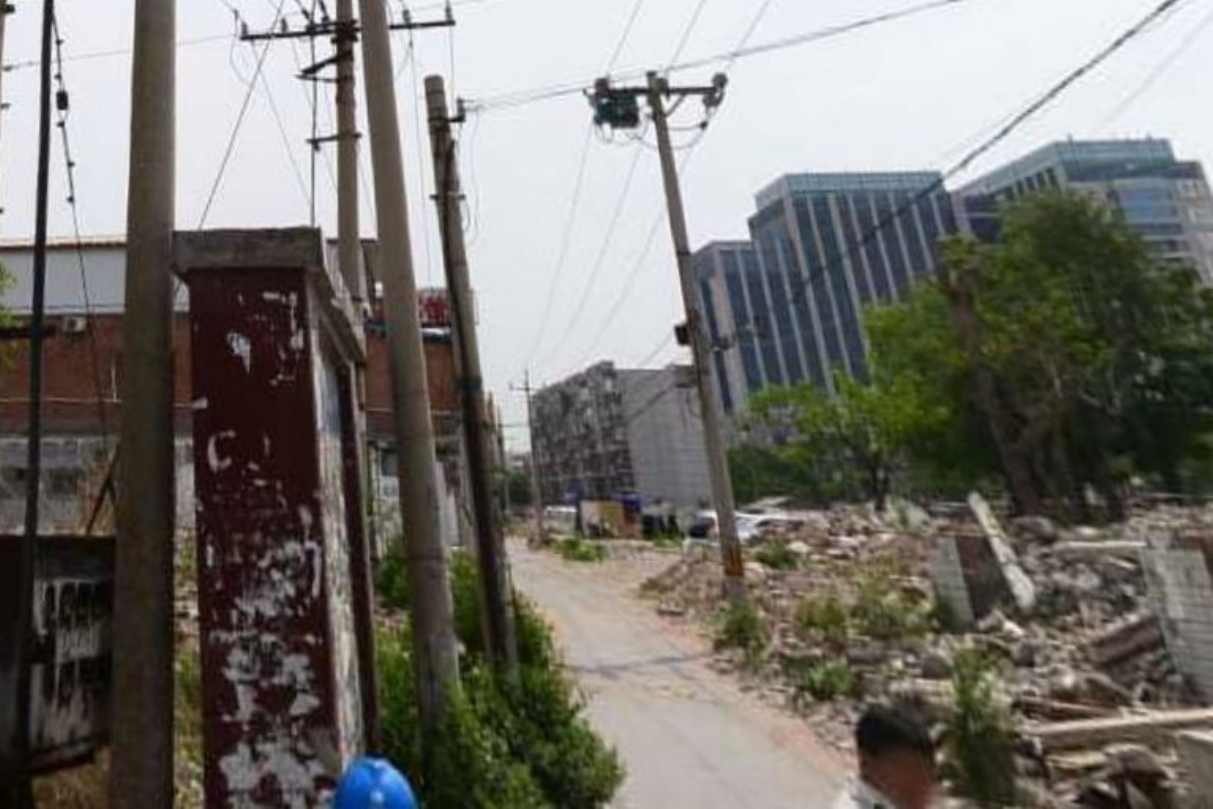}
    }
    \subfigure[The Jijiamiao Village in 2019]{
    \includegraphics[width=0.3\linewidth]{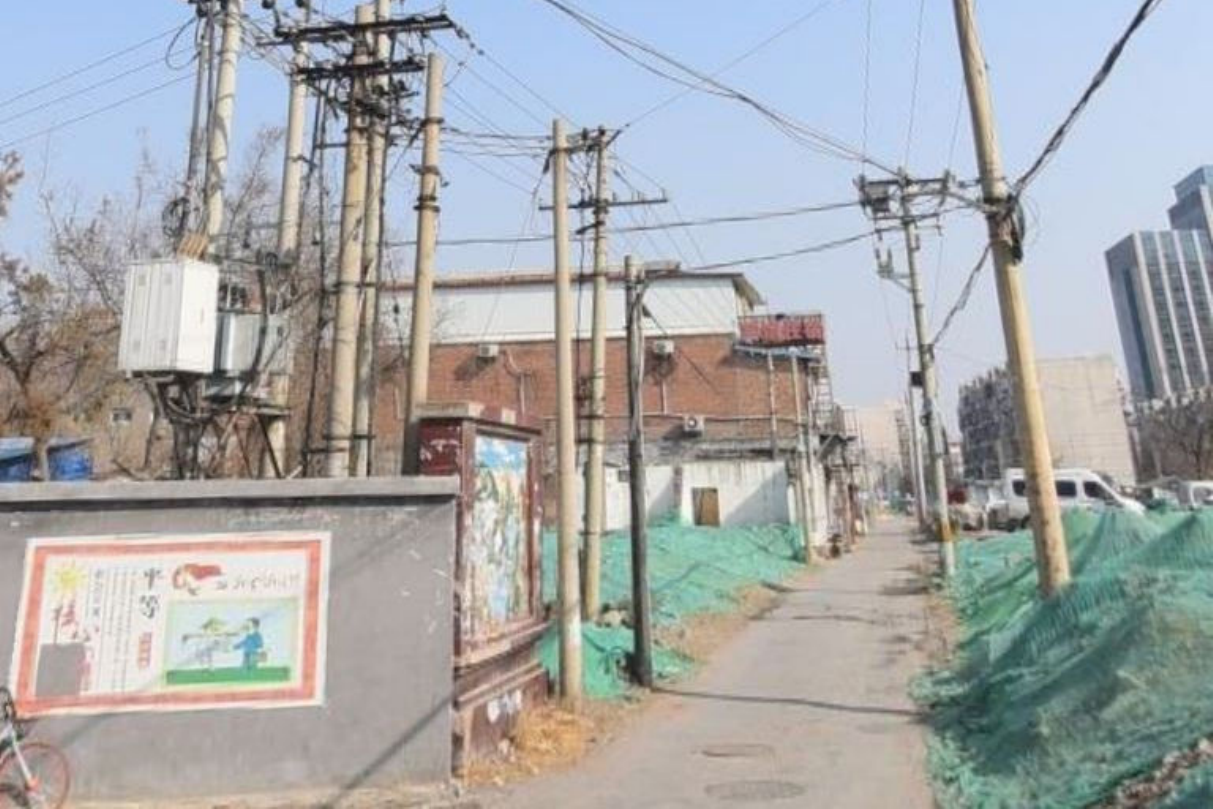}
    }
    \caption{The environment change of the Jijiamiao Village from 2013 to 2019 at street view imagery.}
    \label{fig:beijing_streetview_change}
\end{figure}

\subsection{Evolving Trend with Street View Images in Beijing}

As shown in Figure \ref {fig:beijing_streetview_change}, We show three street view images at the same location in different years. Compared to the chaotic condition of the urban village in 2013, the village had been demolished in 2015. In 2019, new walls were built to improve the city's appearance.

\subsection{Spatial Distribution Analysis with Street View Images in Xi'an}

\begin{figure}[htbp]
\centering
\includegraphics[width=0.4\textwidth]{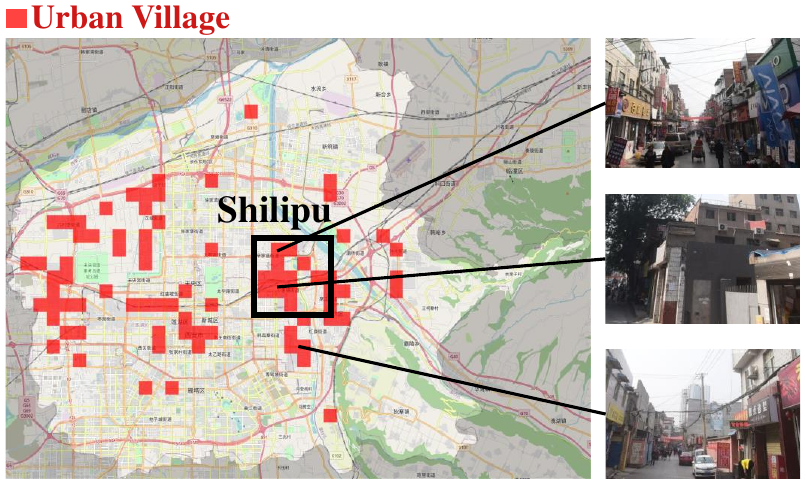}\caption{ Urban village distribution in Xi'an in 2022.}
\label{fig:xian_spatial}
\end{figure}

\begin{figure}[htbp]
\centering
\includegraphics[width=0.4\textwidth]{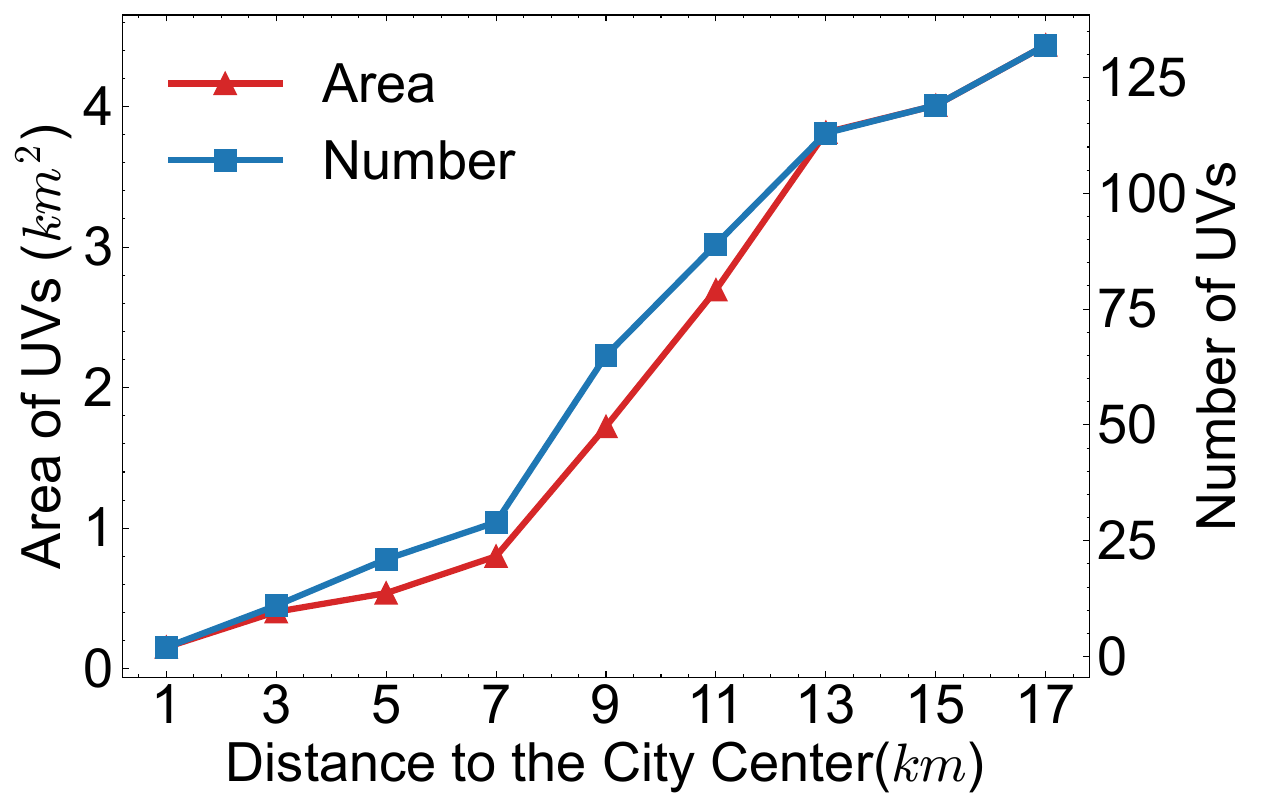}\caption{Urban village (UV) distribution in Xi'an with respect to area and amount in 2022.}
\label{fig:xian_spatial_trend}
\end{figure}
In Figure \ref{fig:xian_spatial}, we visualize the spatial distribution of urban villages within the urban areas of Xi'an, including Lianhu District, Xincheng District, Beilin District, Yanta District, Baqiao District and Weiyang District. In Xi'an, there are few historical reserve blocks.
As shown in the figure, the urban villages within the main urban areas demonstrate a distinct pattern of horizontal distribution. Notably, in the eastern area, a distinct clustering of urban villages is evident in close proximity to Shilipu Village. As for the central area of Xi'an, several urban villages are identified, which might be related to potential misidentification caused by the lower quality of satellite imagery.

Furthermore, we plot distribution curves to visually depict the spatial distribution patterns. In Figure \ref{fig:xian_spatial_trend}, the area and amount of urban villages in 2022 are showcased in relation to their distances from the city center of Xi'an. When the distance from the city center is between 7 to 11 kilometers, there is a steep linear increase in the number and area of urban villages. But at larger distances, the rates of increase become less pronounced. This can be attributed to the "urban-suburban-rural" structure created by the rapid process of urban expansion, where a large number of urban villages are concentrated in the suburbs.

\subsection{Evolving Trend in Xi'an}
\begin{figure}[htbp]
    \centering
    \subfigure[The Yangjia Village in 2013]{
        \includegraphics[width=0.3\linewidth]{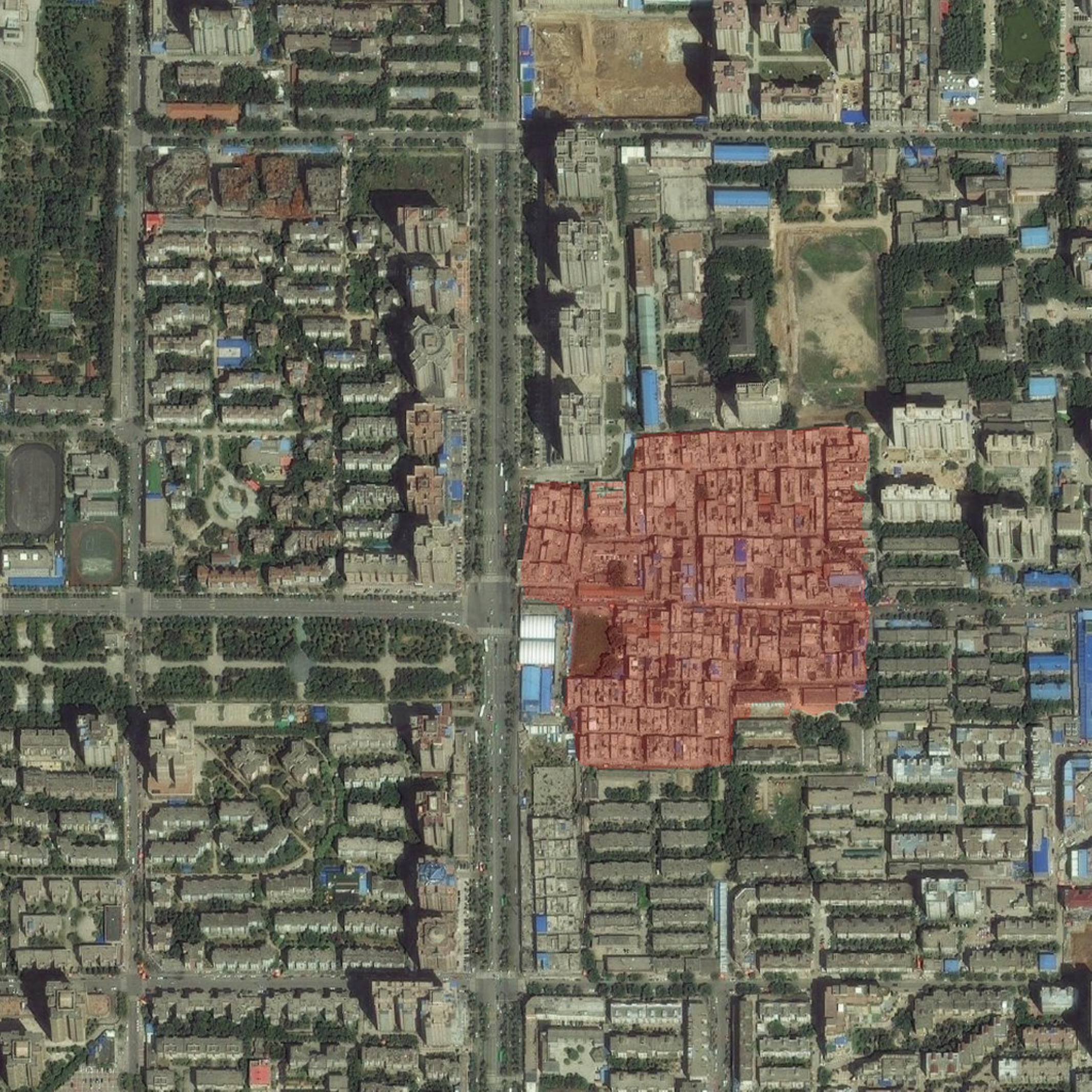}
    }
    \subfigure[The Yangjia Village in 2018]{
    \includegraphics[width=0.3\linewidth]{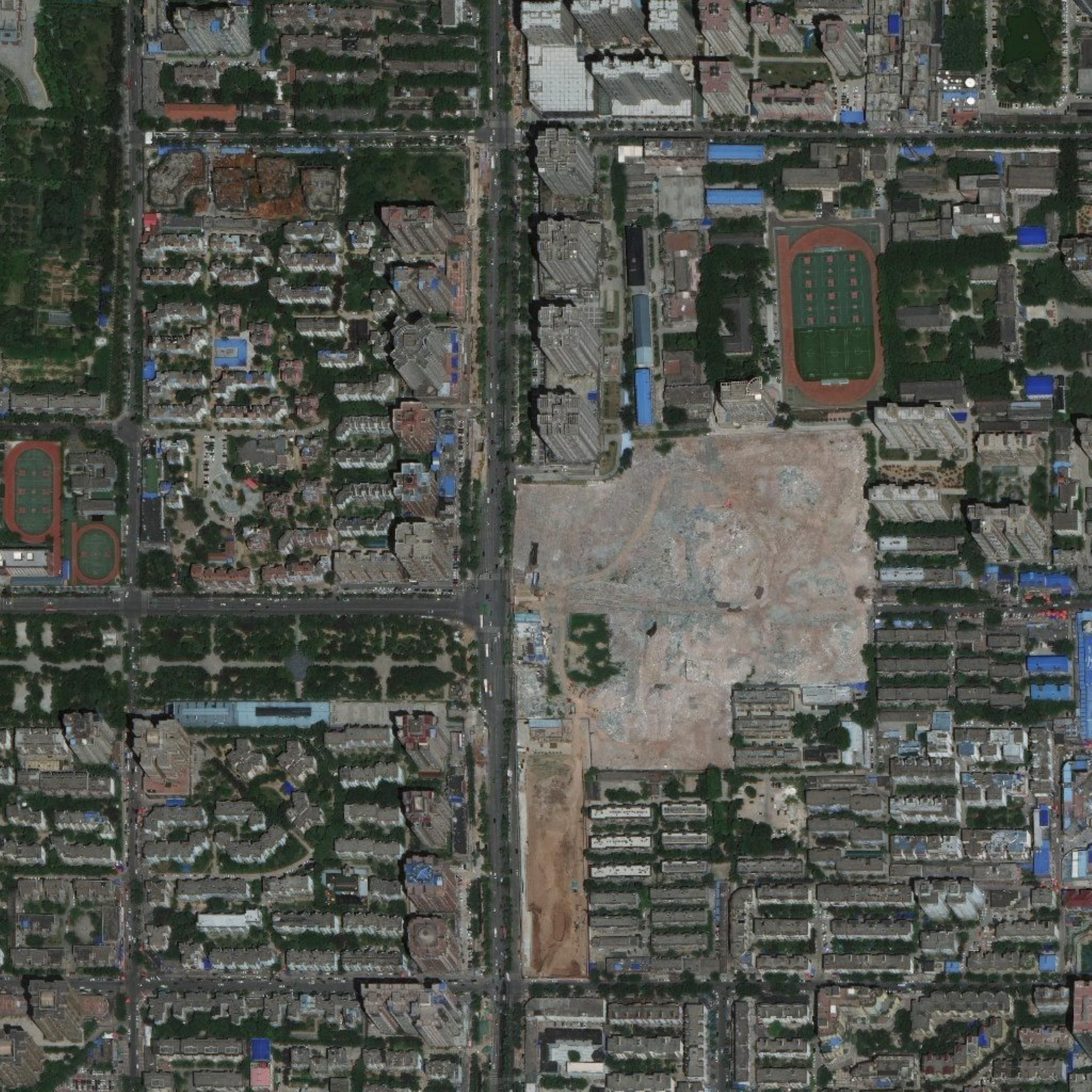}
    }
    \subfigure[The Yangjia Village in 2022]{
    \includegraphics[width=0.3\linewidth]{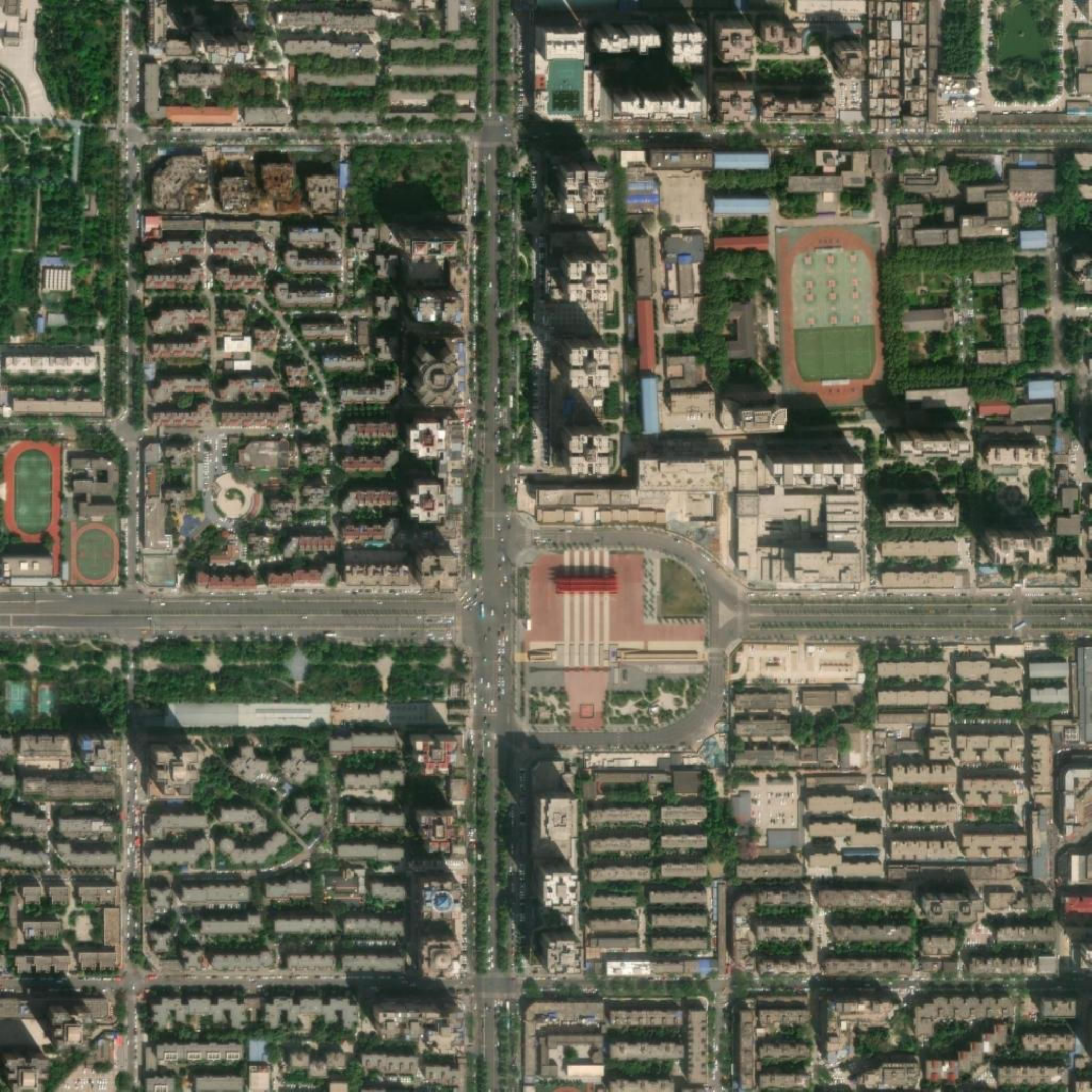}
    }
    \caption{The boundary change of the Yangjia Village from 2013 to 2022. The highlighted red regions denote the areas where urban villages have been identified, signifying a gradual reduction over time.}
    \label{fig:xian_change}
\end{figure}

The Yangjia Village, where the historical ruin of the Mingde Gate is located, became a popular choice for temporary residents due to its affordable rental options. In 2013, the government had planned to take down and renovate the village\footnote{\url{https://www.xa.gov.cn/gk/zcfg/xaszfwj/5d49119d65cbd87465a7851d.html}}. As shown in Figure \ref{fig:xian_change}, the majority of the Yangjia Village was deconstructed in 2018. In 2022, a small part of the Yangjia Village had been transformed into high-rise buildings, while the majority part had been developed into a heritage park.
\begin{figure}[t]
    \centering
    \subfigure[The Yangjia Village in 2014]{
        \includegraphics[width=0.3\linewidth]{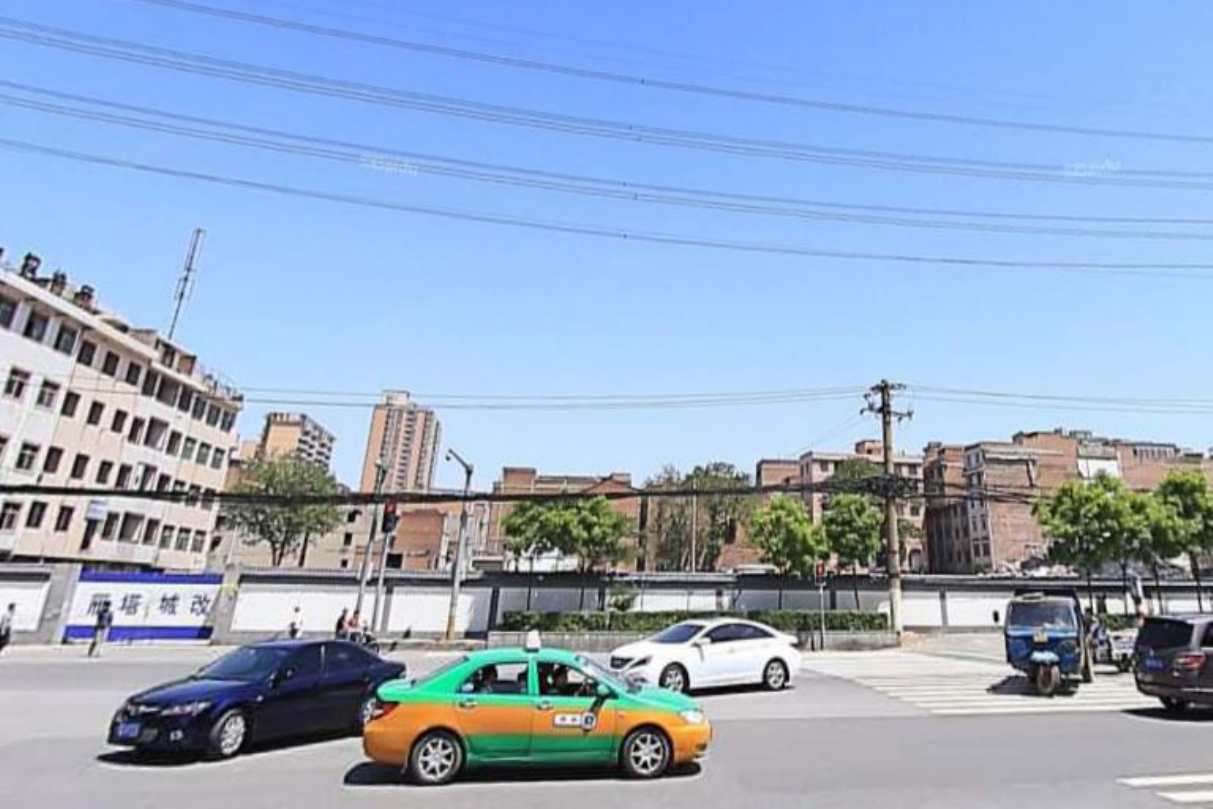}
    }
    \subfigure[The Yangjia Village in 2017]{
    \includegraphics[width=0.3\linewidth]{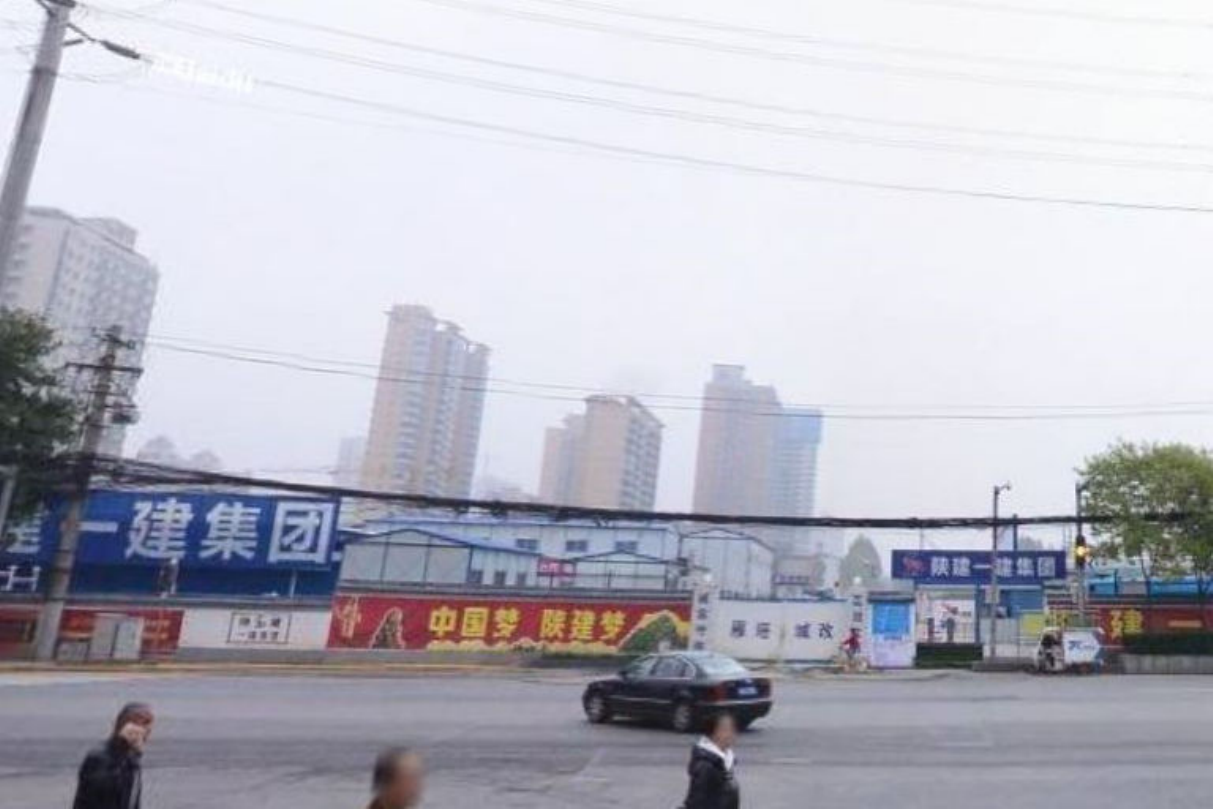}
    }
    \subfigure[The Yangjia Village in 2019]{
    \includegraphics[width=0.3\linewidth]{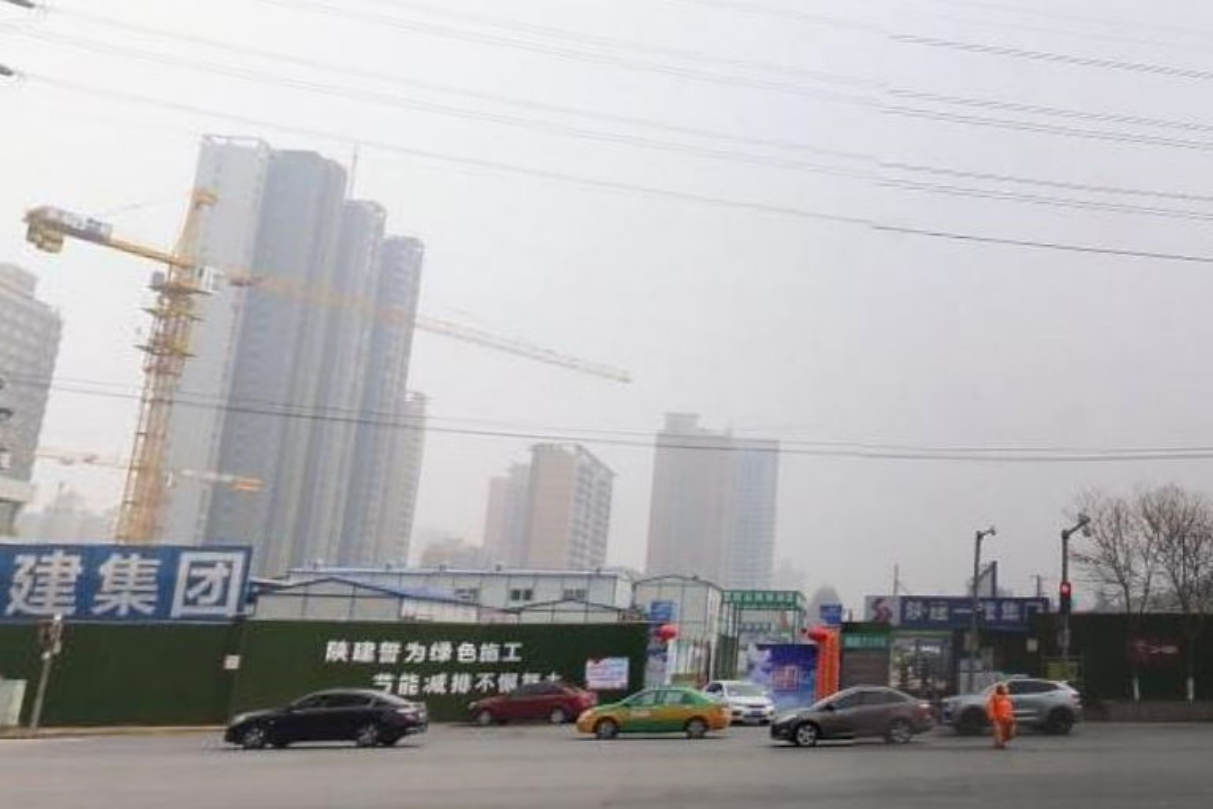}
    }
    \caption{The environment change of the Yangjia Village from 2014 to 2019.}
    \label{fig:xian_streetview_change}
\end{figure}

From the street view images of Yangjia Village displayed in Figure \ref{fig:xian_streetview_change}, it becomes clear that the village started being renovated as early as 2014. By 2019, the emergence of tall buildings became noticeable.

\bibliography{aaai24}

\end{document}